\RequirePackage{fix-cm}
\documentclass[smallextended]{svjour3}          
\smartqed  

\usepackage{graphicx} 
\usepackage{epstopdf} 
\usepackage[normalem]{ulem}

\usepackage{geometry}
\usepackage[export]{adjustbox}
\usepackage[labelfont=bf, labelsep=space]{caption}
\usepackage{subfig}
\usepackage[utf8]{inputenc}
\usepackage{amssymb}
\usepackage{amsmath} 

\usepackage{natbib} 
\usepackage[bookmarks,bookmarksopen,bookmarksdepth=2]{hyperref} 
\hypersetup{colorlinks=true,
			citecolor=blue,
			linkcolor=blue}

\usepackage{tikz} 
\usetikzlibrary{fit,positioning} 
\usepackage{soul} 

\DeclareMathOperator*{\argmin}{arg\,min}

\newcommand{\Muo}{\boldsymbol{\mu}_{0}^\text{(a)}}
\newcommand{\Ro}{\mathbf{R}_{0}^\text{(a)}}
\newcommand{\invRo}{\left(\mathbf{R}_{0}^\text{(a)}\right)^{-1}}

\newcommand{\Wo}{\mathbf{W}_{0}^\text{(a)}}
\newcommand{\invWo}{\left(\mathbf{W}_{0}^{\text{(a)}}\right)^{-1}}
\newcommand{\betaoa}{\beta_{0}^\text{(a)}}

\newcommand{\muo}{\mu_{0}^\text{(f)}}
\newcommand{\ro}{r_{0}^\text{(f)}}
\newcommand{\invro}{\left(r_{0}^\text{(f)}\right)^{-1}}
\newcommand{\wo}{w_{0}^\text{(f)}}

\newcommand{\betaof}{\beta_{0}^\text{(f)}}

\newcommand{\Muk}{\boldsymbol{\mu}_{k}^\text{(a)}}
\newcommand{\Sk}{\mathbf{S}_{k}^\text{(a)}}
\newcommand{\muk}{\mu_{k}^\text{(f)}}
\newcommand{\sk}{s_{k}^\text{(f)}}

\newcommand{\Szu}{\mathbf{S}_{z_u}^\text{(a)}}
\newcommand{\invSzu}{\left(\mathbf{S}_{z_u}^{\text{(a)}}\right)^{-1}}
\newcommand{\szu}{s_{z_u}^\text{(f)}}
\newcommand{\invszu}{\left(s_{z_u}^{\text{(f)}}\right)^{-1}}

\newcommand{\Muzu}{\boldsymbol{\mu}_{z_u}^\text{(a)}}
\newcommand{\muzu}{\mu_{z_u}^\text{(f)}}

\journalname{Comput Stat}

\begin{document}
\title{Non-parametric clustering over user features and latent behavioral functions with dual-view mixture models}
\titlerunning{Non-parametric user clustering with dual-view mixture models}        

\author{Alberto Lumbreras \and
        Julien Velcin  \and\\
        Marie Guégan \and
        Bertrand Jouve
}


\institute{Alberto Lumbreras \and Marie Guégan \at
		   Technicolor\\
           975 Avenue des Champs Blancs, \\35576 Cesson-Sévigné,\\ France\\
           \email{alberto.lumbreras@technicolor.com}\\
           \email{marie.guegan@technicolor.com}
           \and
           Julien Velcin \at
           Laboratoire ERIC, Université de Lyon,\\
           5, avenue Pierre Mendès France, 69676 Bron,\\ France\\
           \email{julien.velcin@univ-lyon2.fr}
		  \and
           Bertrand Jouve \at
           Université de Toulouse; UT2; FRAMESPA/IMT; 5 allée Antonio Machado, 31058 Toulouse, cedex 9\\
           CNRS; FRAMESPA; F-31000 Toulouse\\
           CNRS; IMT; F-31000 Toulouse\\ France\\     
           \email{jouve@univ-tlse2.fr}
}

\date{Received: 27 May 2015 / / Accepted: 15 June 2016}

\maketitle
\begin{abstract}
We present a dual-view mixture model to cluster users based on their features and latent behavioral functions. Every component of the mixture model represents a probability density over a feature view for observed user attributes and a behavior view for latent behavioral functions that are indirectly observed through user actions or behaviors. Our task is to infer the groups of users as well as their latent behavioral functions. We also propose a non-parametric version based on a Dirichlet Process to automatically infer the number of clusters. We test the properties and performance of the model on a synthetic dataset that represents the participation of users in the threads of an online forum. Experiments show that dual-view models outperform single-view ones when one of the views lacks information.
\keywords{Multi-view clustering \and Model-based clustering \and Dirichlet Process (DP) \and Chinese Restaurant Process (CRP) }
\end{abstract}
\section{Introduction}\label{sec:introduction}
We consider the problem of clustering users over both their observed features and their latent behavioral functions. The goal is to infer the behavioral function of each user and a clustering of users that takes into account both their features and their behavioral functions. In our model, users in the same cluster are assumed to have similar features \textit{and} behavioral functions, and thus the inference of the clusters depends on the inference of the behavioral functions, and vice versa.

Latent behavioral functions are used to model individual behaviors such as pairwise choices over products, reactions to medical treatments or user activities in online discussions. The inclusion of latent behavioral functions in clustering methods has several potential applications. On the one hand, it allows richer clusterings in settings where users, besides being described through feature vectors, also perform some observable actions that can be modelled as the output of a latent function. On the other hand, by assuming that users in the same cluster share similar latent functions, it may leverage inference of these functions. In the case, for instance, of recommender systems, this may be used to alleviate the \textit{cold-start problem} ---the fact that we cannot infer user preferences until they have interacted with the system for a while--- if we have a set of features describing user profiles. In that context, a user with low activity will be given the same cluster as those users with similar features. Then, the inference of its behavioral function (e.g.: its preferences) will be based on the behavioral functions of the users in the same cluster.

One of the difficulties in dealing with latent behavioral functions is that, since these functions are latent, they are not representable in a feature-like way and therefore traditional clustering methods are not directly applicable.
Our approach is to think of features and behaviors as two different \textit{views} or representations of users, and to find the partition that is most consensual between the different views. In this sense, this is a multi-view clustering problem \citep{SteffenandTobiasScheffer}. However, the clustering in one of the views depends on latent variables that need to be inferred. In this regard, it has similarities to Latent Dirichlet Allocation when used for clustering (e.g., cluster together documents that belong to the same latent topics).

Multi-view clustering ranges from \cite{Kumar2011}, which finds a consensual partition by co-training \citep{Mitchell1998}, to \cite{Greene2009a}, which proposes a two-step multi-view clustering that allows both consensual groups and groups that only exist in one of the views. In \cite{Niu2012}, the multi-view approach is presented as the problem of finding multiple cluster solutions for a single description of features.

In this paper, we extend the idea of multi-view clustering to deal with cases where one of the views comprises latent functions that are only indirectly observed through their outputs. The proposed method consists of a dual-view mixture model where every component represents two probability densities: one over the space of features and the other over the space of latent behavioral functions. The model allows to infer both the clusters and the latent functions.  Moreover, the inference of the latent functions allows to make predictions on future user behaviors. The main assumption is that users in the same cluster share both similar features and similar latent functions and that users with similar features and behaviors are in the same cluster. Under this assumption, we show that this dual-view model requires less examples than single-view models to make good inferences.

The idea of using similarities in one view to enhance inference in the other view is not new. In bioinformatics, \cite{Eisen1998} found evidence suggesting that genes with similar DNA microarray expression data may have similar functions. \cite{Brown2000} exploit this evidence to train a Support Vector Machine (SVM) for each functional class and predict whether an unclassified gene belongs to that class given its expression data as an input. \cite{Pavlidis2002} extend the method of Brown et al. by also exploiting evidence that similar phylogenetic profiles between genes suggested a same functional class as well \citep{Pellegrini1999a}. Pavlidis et al. propose a multi-view SVM method that uses both types of gene data as input.
 
More recently, \cite{Cheng2014} applied multi-view techniques to predict user labels in social networks such as LinkedIn (e.g., engineer, professor) or IMdB (e.g., directors, actors). Their method lies in the maximization of an objective function with a co-regularization that penalizes predictions of different labels for users that are similar either in terms of profile features or in terms of graph connectivity.

In the context of preference learning, \cite{Bonilla2010} also work with the assumption that similar users may have similar preferences, and models this assumption via a Gaussian Process prior over user utility functions. This prior favors utility functions that account for user similarity and item similarity. To alleviate the computational cost of this model, \cite{Abbasnejad2013a} propose an infinite mixture of Gaussian Processes that generates utility functions for groups of users assuming that users in each community share one single utility function. 
The main difference between our model and \cite{Abbasnejad2013a} in that their model clusters users only based on their utility functions, while ours considers user features as well. In short, ours is a multi-view clustering model that also infers latent behavioral functions, while theirs is a single-view model focused on the inference of latent functions.

The paper is structured as follows: first, we briefly recall mixture models. Second, we present our model as an extension of classic mixture models. The description of the model ends up with a generalization to an infinite number of clusters, which makes the model non-parametric. We finally describe an application to cluster users in online forums and end the paper with experiments on synthetic data to demonstrate the properties of the model.

\section{Model description}\label{sec:modeldescription}
In this section, we introduce our model through three sequential steps. First, we start with a simple mixture model. Second, we extend the mixture model to build a \textit{dual-view} mixture model. And third, we extend the \textit{dual-view} mixture model so that the number of clusters is automatically inferred.

\subsection{Mixture models}\label{sec:mixturemodels}
When a set of observations $x_1,x_2,...x_n$ cannot be properly fitted by a single distribution, we may get a better fit by considering that different subsets of observations come from different component distributions. Then, instead of a unique set of parameters $\boldsymbol{\theta}$ of a single distribution, we need to infer $K$ sets of parameters $\boldsymbol{\theta}_1,...,\boldsymbol{\theta}_K$ of $K$ components and the assignments $z_1,...,z_n$ of individual observations to one of these components. The model can be expressed as follows:
\begin{align}
x_i | z_i, \boldsymbol{\theta}_{z_i} &\sim F(\boldsymbol{\theta}_{z_i})\notag\\
z_i &\sim \text{Discrete}(\boldsymbol{\pi})
\end{align}
where $\boldsymbol{\pi} = (\pi_1,...\pi_K)$ contains the probability of belonging to each component and $F$ is the likelihood function over the observations. In Bayesian settings it is common to add priors over these parameters, resulting in a model such as:
\begin{align}
x_i | z_i, \boldsymbol{\theta}_{z_i} &\sim F(\boldsymbol{\theta}_{z_i})\notag\\
\boldsymbol{\theta}_j &\sim G_0\notag\\
z_i | \boldsymbol{\pi} &\sim \text{Discrete}(\boldsymbol{\pi})\notag\\
\boldsymbol{\pi} &\sim \text{Dirichlet}(\alpha)
\end{align}
where $G_0$ is the \emph{base distribution} and $\alpha$ the \textit{concentration parameter}. Mixture models are mostly used for \textit{density estimation}. Nonetheless, inference over $\mathbf{z}$ allows to use them as \textit{clustering} methods. In this case, every component is often associated to a cluster.

\subsection{Dual-view mixture model}\label{sec:dual-view}
In this section, we present an extension of mixture models to account both for features and latent behavioral functions. We denote by \textit{behavioral functions} any function which, if known, can be used to predict the behavior of a user in a given situation. In the context of preference learning \citep{Bonilla2010,Abbasnejad2013a} or recommender systems \citep{Cheung2004}, behavioral functions may indicate hidden preference patterns, such as utility functions over the items,  and the observed behavior may be a list of pairwise choices or ratings. In the context of online forums, behavioral functions may indicate the reaction of a user to a certain type of post and the observed behavior may be the set of replies to different posts. In general, behavioral functions are linked to observed behaviors through a likelihood function $p(y | f)$ where $y$ represents an observation and $f$ the latent behavioral function. 

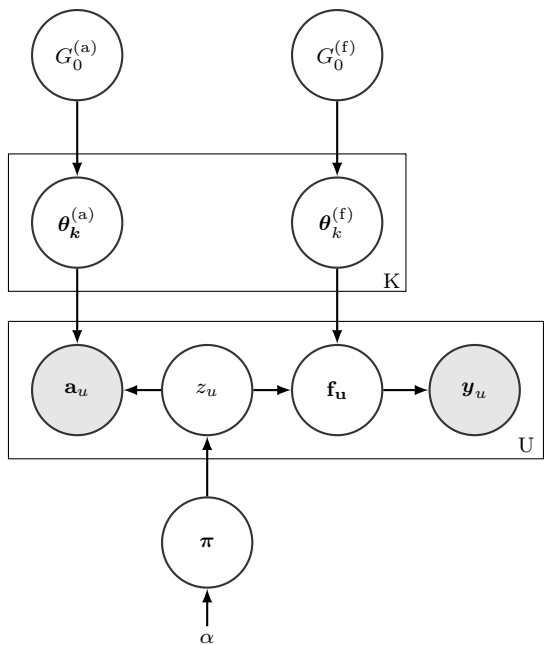
\begin{figure}
\center
 \scalebox{1}{
\begin{tikzpicture}
\tikzstyle{main}=[circle, minimum size = 12mm, thick, draw =black!80, node distance = 8mm]
\tikzstyle{connect}=[-latex, thick]
\tikzstyle{box}=[rectangle, draw=black!100]

  \node (alpha) {$\alpha$}; 
  \node [main](pi) [above=0.5cm of alpha] {$\boldsymbol{\pi}$}; 
  \node [main](z) [above=of pi] {$z_u$};

  \node[main, fill = black!10] (a_u) [left=0.5cm of z] {$\mathbf{a}_u$}; 
  \node[main] (theta_ar) [above=1cm of a_u] {$\boldsymbol{\theta_k^{\text{(a)}}}$};   

  \node[main] (f_u) [right=0.5cm of z] {$\mathbf{f_u}$}; 
  \node[main] (theta_fr) [above=1cm of f_u] {$\boldsymbol{\theta}_k^{\text{(f)}}$};   
  \node[main, fill = black!10] (y) [right=0.6cm of f_u] {$\boldsymbol{y}_u$};   
   
  \node [main](G_a0) [above=1cm of theta_ar] {$G_0^{\text{(a)}}$};   
  \node [main](G_f0) [above=1cm of theta_fr] {$G_0^{\text{(f)}}$};   

  \path 
           (alpha) edge [connect] (pi)
           (pi) edge [connect] (z)
           (z) edge [connect] (f_u)
           (z) edge [connect] (a_u)
           
           (G_a0) edge [connect] (theta_ar)
           (theta_ar) edge [connect] (a_u)

           (G_f0) edge [connect] (theta_fr)
           (theta_fr) edge [connect] (f_u)
           (f_u) edge [connect] (y);

  \node[rectangle, inner sep=0mm, fit= (a_u) (z) (f_u), label=below right:U, yshift=1mm, xshift=34mm] {};
  \node[rectangle, inner sep=3mm,draw=black!100, fit= (z) (f_u) (a_u) (y)] {};  

  \node[rectangle, inner sep=1mm, fit= (theta_ar) (theta_fr), label=above left:K, yshift=-17mm, xshift=34mm] {};
  \node[rectangle, inner sep=3mm, draw=black!100, fit= (theta_fr) (theta_ar)] {};  
\end{tikzpicture}
}
\caption{Graphical model of the generative process for $U$ users and $K$ clusters. Shaded circles represent observations and white circles represent latent variables. Views are connected through the latent assignments $\mathbf{z}$. A user $u$ draws a feature vector $\mathbf{a}_u$ and a behavior $\mathbf{f}_u$ from the cluster indicated by $z_u$ (the $u$-th element of $\mathbf{z}$).}
\label{fig:general}
\end{figure}

Let $\mathbf{a}_u$ be the set of (observed) features of user $u$. Let $f_u$ be a (latent) function of user $u$. Let $y_u$ be the (observed) outcome of $f_u$. By slightly adapting the notation from last section we can describe the variables of our dual model as follows: 

\begin{align}
\mathbf{a}_u | z_u, \boldsymbol{\theta}_{z_u}^{\text{(a)}} &\sim F^{\text{(a)}}(\boldsymbol{\theta}_{z_u}^{\text{(a)}})\notag\\
f_u | z_u, \boldsymbol{\theta}_{z_u}^{\text{(f)}} &\sim F^{\text{(f)}}(\boldsymbol{\theta}_{z_u}^{\text{(f)}})\notag\\
y_u | f_u &\sim p(y_u | f_u)\notag\\
\boldsymbol{\theta}_j^{\text{(a)}} &\sim G_0^{\text{(a)}}\notag\\
\boldsymbol{\theta}_j^{\text{(f)}} &\sim G_0^{\text{(f)}}\notag\\
z_u | \boldsymbol{\pi} &\sim \text{Discrete}(\boldsymbol{\pi})\notag\\
\boldsymbol{\pi} &\sim \text{Dirichlet}(\alpha)
\end{align}
where we use the superindex $(a)$ for elements in the \textit{feature view} and the superindex $(f)$ for elements in the latent functions view, henceforth \textit{behavior view}. Otherwise, the structures are  similar except for $y_u$, which represents the set of observed behaviors for user $u$. The corresponding Probabilistic Graphical Model is shown in Figure \ref{fig:general}.

Every component has two distributions: one for features and one for latent behavioral functions. Latent behavioral functions are not directly observable, but they may be inferred through some observations if we have a likelihood function of observations given the latent functions.   

The model can also be expressed in terms of a generative process:
\begin{itemize}
\item For every component $k$:
   \begin{itemize}
   \item Draw feature and function parameters from their base distributions $\boldsymbol{\theta}_k^{\text{(a)}} \sim G_0^{\text{(a)}}$ and $ \boldsymbol{\theta}_k^{\text{(f)}} \sim G_0^{\text{(f)}}$.
   \end{itemize}
\item Draw the mixture proportions $\boldsymbol{\pi} \sim \text{Dirichlet}(\alpha)$.
\item For every user $u$:
   \begin{itemize}
   \item Draw a component assignment $z_u \sim \text{Multinomial}(\boldsymbol{\pi})$.
   \item Draw user features  $\mathbf{a}_u \sim F^{\text{(a)}}(\boldsymbol{\theta}_{z_u}^{\text{(a)}})$. 
   \item Draw a user latent function $f_u \sim F^{\text{(f)}}(\boldsymbol{\theta}_{z_u}^{\text{(f)}})$.
   \item Draw a set of observed behaviors $y_u \sim p(y_u | f_u)$.
   \end{itemize}
\end{itemize}

Left and right branches of Figure \ref{fig:general}  correspond to the \textit{feature view} and the \textit{behavior view}, respectively. Note that every component contains two sets of parameters, one for features and one for behaviors, so that the two views can be generated from the same component. This encodes our prior assumption that users who belong to the same cluster should be similar in both views.

Given the user assignments $\mathbf{z}$, variables in one view are conditionally independent from variables in the other view. That means their inferences can be considered separately. However, inference of $\mathbf{z}$ uses information from both views. The conditional probability of $\mathbf{z}$ given all the other variables is proportional to the product of its prior and the likelihood of both views:
\begin{align}
p(\mathbf{z} | \cdot) 
\propto
p(\mathbf{z} | \boldsymbol{\pi}) 
p(\mathbf{a} | \boldsymbol{\theta^{\text{(a)}}}, \mathbf{z}) 
p(\mathbf{f} | \boldsymbol{\theta^{\text{(f)}}}, \mathbf{z})
\label{eq:z_posterior}
\end{align}
\sloppy
The information given by each view is conveyed through the likelihood factors $p(\mathbf{a} | \boldsymbol{\theta^{\text{(a)}}}, \mathbf{z})$ and $p(\mathbf{f} | \boldsymbol{\theta^{\text{(f)}}}, \mathbf{z})$. The ratio between the conditional probability of a partition $\mathbf{z}$ and the conditional probability of a partition $\mathbf{z'}$ is:
\begin{align}
\frac
{p(\mathbf{z}| \cdot)}
{p(\mathbf{z'}| \cdot)}
= 
\frac{
p(\mathbf{z} | \boldsymbol{\pi})
}{
p(\mathbf{z'} | \boldsymbol{\pi})
}
\frac{
p(\mathbf{a} | \boldsymbol{\theta^{\text{(a)}}}, \mathbf{z})
}{
{p(\mathbf{a} | \boldsymbol{\theta^{\text{(a)}}}, \mathbf{z'})}
}
\frac{
p(\mathbf{f} | \boldsymbol{\theta^{\text{(f)}}}, \mathbf{z})
}{
{p(\mathbf{f} | \boldsymbol{\theta^{\text{(f)}}}, \mathbf{z'})}
}
\label{eq:z_posterior_ratio}
\end{align}
\fussy
where the contribution of each view depends on how much more likely $\mathbf{z}$ is over the other assignments in that view. An extreme case would be a uniform likelihood in one of the views, meaning that all partitions $\mathbf{z}$ are equally likely. In that case, the other view leads the inference. 

The two views provide reciprocal feedback to each other through $\mathbf{z}$. This means that if one view is more confident about a given $\mathbf{z}$, it will not only have more influence on $\mathbf{z}$ but also it will force the other view to re-consider its beliefs and adapt its latent parameters to fit the suggested $\mathbf{z}$. 

Note also that inference of latent behavioral functions may be used for prediction of future behaviors.

\subsection{Infinite number of clusters}\label{sec:CRP}
So far, we have considered the number of components $K$ to be known. Nonetheless, if we let $K \rightarrow \infty$ and marginalize over the mixture weights $\boldsymbol{\pi}$, it becomes a non-parametric model with a Dirichlet Process (DP) based on a Chinese Restaurant Process (CRP) prior over the user assignments, which automatically infers the number of \textit{active} of components (see the Appendix for the full derivation). Since we integrate out $\boldsymbol{\pi}$, user assignments are not independent anymore.
Instead, the probability of a user $u$ to be assigned to a non-empty (active) component $k$, given the assignments of all other users $\mathbf{z}_{-u}$, is:
\begin{align}
	p(z_u = k | \mathbf{z}_{-u}) &\propto n_k \qquad \text{ for } k = 1,...,c
\end{align}
where $c$ denotes the number of non-empty components and $n_k$ is the number of users already assigned to the $k$-th component. The probability of assigning a user $u$ to an empty (non-active) component, that would be labelled as $c+1$, is: 

\begin{align}
	p(z_u = k | \mathbf{z}_{-u}) &\propto \alpha \qquad  \text{ for } k=c+1
\end{align}
These two equations reflect a generative process that assigns users to clusters in \textit{a rich get richer} manner. The more users in a component, the more attractive this component becomes. Empty components also have a chance of being filled. 
Despite the appearance of these equations, the idea behind the inference of $\mathbf{z}$ remains the same. The only differences between the finite and the infinite cases are the number of components and the probabilities to be assigned to each component.


\section{Application to role detection in online forums}\label{sec:forums}

The above model provides a general framework that can be adapted to many scenarios. In this section, we apply our model to the clustering of users in online forums. Clustering users in online communities may be used for latent \textit{role detection}. Although clustering based on user features may provide interesting insights, we think that the notion of \textit{role} should include information that allows to predict behaviors. After all, this is what roles are useful for. We expect that a person holding a role behaves according to their role.

Let us specify the two views of our model for this scenario, given $U$ users who participate in a forum composed of $T$ discussion threads. For the feature view, we describe every user through a feature vector $\mathbf{a}_u=(a_{u1}, a_{u2},...,a_{uD})^T$ that will typically contain features such as centrality metrics or number of posts. For the behavior view, we define a latent behavioral function that we call \textit{catalytic power} and denote by $b_u$, which represents the ability of a user to promote long discussions; we refer to $\mathbf{b}$ as the vector of all user catalytic powers. Let the \textit{participation vector} of the discussion thread $\mathbf{p}_t = (p_{1t},...,p_{Ut})^T$ be a binary array indicating which users participated among the first $m$ posts of the thread $t$. Assuming  that the dynamic of a discussion is strongly conditioned by the first participants, we model the final length of a thread $y_t$:
\begin{equation*}
y_t \sim \mathcal{N}(\mathbf{p}_t^T\mathbf{b}, s_{\text{y}}^{-1})
\end{equation*} 
where $\mathbf{p}_t^T\mathbf{b}$ is the cumulated catalytic power of users who participated in its first $m$ posts and $s_{\text{y}}$ represents the precision (inverse of the variance) of the unknown level of noise.

If the assumption that there exist groups of users sharing similar features and similar catalytic power holds, then our model will not only find a clustering based on features and behaviors (catalytic powers), but it will also exploit feature information to infer catalytic powers and, vice versa, the inferred catalytic powers will be treated by the model as an extra feature dimension.

Note that, unlike the model presented in Figure \ref{fig:general}, the observed output $y_t$ is common to all users who participated in the first $m$ posts of thread $t$. Moreover, $y_t$ depends on the observed participations $\mathbf{p_t}$. We also defined the noise factor $s_{\text{y}}$ which was not explicitly present in the general model. The graphical model would be similar to that of Figure~\ref{fig:general} but with the thread lengths $\mathbf{y}$, the participation matrix $\mathbf{P}$ and the noise $s_{\text{y}}$ out of the users plate. In the remaining of this section we provide more details about the components of the two views.

\subsection{Feature view}\label{sec:forums_features}
In this section we specify the component parameters $\boldsymbol{\theta}_k^{\text{(a)}}$ and the base distribution $G_0^{\text{(a)}}$ of the feature view. Our choice of priors follows that of the Infinite Gaussian Mixture Model (IGMM) as described by  \cite{Rasmussen2000a} and  \cite{Gorur2010}.

The feature vectors are assumed to come from a mixture of Gaussian distributions: 
\begin{align}
\boldsymbol{a}_u &\sim \mathcal{N}\left(\Muzu, \invSzu\right)
\end{align}
where the mean $\Muzu$ and the precision $\Szu$ are component parameters common to all users assigned to the same component. The component parameters are given Normal and Wishart priors: 
\begin{align}
\Muk &\sim  \mathcal{N}\left(\Muo, \invRo\right)\\
\Sk &\sim  \mathcal{W}\left( \betaoa, \left( \betaoa \Wo \right)^{-1}\right) 
\label{eq:wishart_sar}
\end{align}
where the mean $\Muo$, the precision $\Ro$, the covariance $\Wo$, and the degrees of freedom $\betaoa$ are hyperparameters common to all components. The hyperparameters themselves are given non-informative priors centered at the observed data
\begin{align}
\Muo &\sim \mathcal{N}(\boldsymbol{\mu_a, \Sigma_a}) \\
\Ro &\sim \mathcal{W}(D, (D \boldsymbol{\Sigma_a})^{-1})\\
\Wo &\sim \mathcal{W}(D, \frac{1}{D} \boldsymbol{\Sigma_a})\\
\frac{1}{\betaoa - D + 1} &\sim \mathcal{G}(1, \frac{1}{D}) \label{eq:gamma_ba0}
\end{align}
where $\boldsymbol{\mu_a}$ and $\boldsymbol{\Sigma_a}$ are the mean and covariance of all the features vectors and $\mathcal{G}$ is the Gamma distribution. 
Note that the expectation of a random matrix drawn from a Wishart distribution $X \sim \mathcal{W}(\upsilon, W)$ is $\mathbb{E}[\mathbf{X}] = \mathbf{\upsilon W}$. Our parametrization of the Gamma distribution corresponds to a one-dimensional Wishart. Its density function is therefore given by $\mathcal{G}(\alpha, \beta) \propto
x^{\alpha/2-1} \exp(-\frac{x}{2\beta})$
and the expectation of a random scalar drawn from a Gamma distribution $x \sim \mathcal{G}(\alpha, \beta)$ is ${\mathbb{E}[x] = \alpha\beta}$. 

As pointed out in \cite{Gorur2010}, this choice of hyperparameters, which is equivalent to scaling the data and using unit priors, makes the model invariant to translations, rotations, and rescaling of the data. 
Conjugate priors are chosen whenever possible to make the posteriors analytically accessible. As for $\betaoa$, the prior in Equation \ref{eq:gamma_ba0} guarantees that the degrees of freedom in the Wishart distribution in Equation~\ref{eq:wishart_sar} are greater than or equal to $D-1$. The density $p(\betaoa)$ is obtained by a simple transformation of variables (see Appendix).

\subsection{Behavior view}\label{sec:forums_behaviors}
In this section we specify the component parameters $\boldsymbol{\theta}_k^{\text{(f)}}$ and the base distribution $G_0^{\text{(f)}}$ of the behavior view. Our choice corresponds to a Bayesian linear regression where coefficients are drawn not from a single multivariate Gaussian but from a \emph{mixture} of one-dimensional Gaussians.

The thread lengths are assumed to come from a Gaussian distribution whose mean is determined by the catalytic power of users who participated in the first posts and whose variance represents the unknown level of noise:
\begin{align}
y_t &\sim \mathcal{N}(\mathbf{p}_t^T \mathbf{b}, s_{\text{y}}^{-1})
\end{align}
where the precision $s_{\text{y}}$ is given a Gamma prior centered at the sample precision $\sigma_0^{-2}$:
\begin{align}
s_{\text{y}} \sim \mathcal{G}(1,\sigma_{\text{0}}^{-2})
\end{align}
The power coefficients $b_u$ come from a mixture of Gaussians:
\begin{align}
b_u &\sim \mathcal{N}\left(\muzu, \invszu\right)
\end{align}
where the mean $\muzu$ and the precision $\szu$ are component parameters common to all users assigned to the same component $z_u$. The component parameters are given Normal and Gamma priors:
\begin{align}
\muk &\sim  \mathcal{N}\left(\muo, \invro \right)\\
\sk &\sim  \mathcal{G}\left( \betaof, \left(\betaof\wo\right)^{-1}\right) 
\end{align}
where the mean $\muo$, the precision $\ro$, the variance $\wo$, and the degrees of freedom $\betaof$ are hyperparameters common to all components. Because the coefficients are not observed, we cannot center the hyperparameters in the data as we did in the feature view. Instead, we use their Maximum Likelihood Estimates, computed as $\mathbf{\mathbf{\hat{b}}} = (\mathbf{P}\mathbf{P^T} + \lambda\mathbf{I})^{-1} \mathbf{P^T y}$, with a regularization parameter $\lambda=0.01$, where $\mathbf{P}$ is the participation matrix $\mathbf{P} = \{\mathbf{p_1},...,\mathbf{p_T}\}$. Then the hyperparameters are given non-informative priors centered at $\mathbf{\hat{b}}$:
\begin{align}
\muo &\sim \mathcal{N}(\mu_{\hat{b}}, \sigma_{\hat{b}}^2) \\
\ro &\sim \mathcal{G}(1, \sigma_{\hat{b}}^{-2})\\
\wo  &\sim \mathcal{G}(1, \sigma_{\hat{b}}^{2})\\
\frac{1}{\betaof}&\sim \mathcal{G}(1, 1)
\end{align}
where $\mu_{\hat{b}}$ and $\sigma_{\hat{b}}^2$ are the mean and the variance of the Maximum Likelihood Estimators $\mathbf{\hat{b}}$ of the coefficients. Note that this choice is data-driven and, at the same time, the risk of overfitting is reduced since hyperparameters are high in the model hierarchy.

\subsection{Shared parameters}\label{sec:sharedparams}
As for the common variables $\mathbf{z}$ and $\alpha$, $\mathbf{z}$ is given a Chinese Restaurant Process prior:
\begin{align}
\label{eq:prior_z_CRP}
p(z_u = k | \mathbf{z}_{-u}, \alpha) \propto 
\begin{cases}
n_{k} & \text{for } k=1,...,c\\
\alpha & \text{for } k=c+1
\end{cases}
\end{align}
where $c$ denotes the number of non-empty components before the assignment of $z_u$ and $n_k$ is the number of users already assigned to the $k$-th component. The concentration parameter $\alpha$ is given a vague inverse gamma prior:
\begin{align*}
\frac{1}{\alpha} \sim \mathcal{G}(1,1)
\end{align*}

\section{Inference}\label{sec:inference}
The latent parameters of our model can be inferred by using Gibbs sampling, and sequentially taking samples of every variable given the others. Conditional distributions are detailed in the appendices. A single iteration of the Gibbs sampler goes as follows: 

\begin{itemize}
\item Sample component parameters $\Sk, \Muk$ conditional on the indicators $\mathbf{z}$ and all the other variables of the two views.
\item Sample hyperparameters $\Muo$, $\Ro$, $\Wo$, $\betaoa$ conditional on  the indicators $\mathbf{z}$ and all the other variables of the two views.
\item Sample component parameters $\sk, \muk$ conditional on the indicators $\mathbf{z}$ and all the other variables of the two views.
\item Sample hyperparameters $\muo, \ro, \wo, \betaof$ conditional on the indicators $\mathbf{z}$ and all the other variables of the two views.
\item Sample coefficients $\mathbf{b}$ conditional on the indicators $\mathbf{z}$ and all the other variables of the two views.
\item Sample $s_{\text{y}}$ conditional on the indicators $\mathbf{z}$ and all the other variables of the two views.
\item Sample indicators $\mathbf{z}$ conditional on all the variables of the two views.
\end{itemize} 

Since we use conjugate priors for almost all the variables, their conditional probabilities given all the other variables are analytically accessible. The degrees of freedom $\betaoa, \betaof$ and the concentration parameter $\alpha$ can be sampled by Adaptive Rejection Sampling \citep{Gilks1992}, which exploits the log-concavity of  $p(\log \beta_0^{\text{(a)}} | \cdot)$,  $p(\log \beta_0^{\text{(f)}} | \cdot)$ and $p(\log \alpha | \cdot)$ (see Appendix). As for the sampling of $\mathbf{z}$, the conditional probability of assigning user $u$ to an active component $k$ is proportional to the prior times the likelihoods:

\begin{align}
p(z_u = k | \mathbf{z_{-u}}, \alpha, \cdot)
\propto 
n_{k}
   p(\mathbf{a}_{u} |\Muk, \Sk)    
   p({b_u} | \muk, \sk) \qquad \text{for } k=1,...,c
\end{align}
and for the conditional probability of assigning $z_u$ to a non-active component:
\begin{align}
p(z_u = k | \mathbf{z}_{-u}, \alpha, \cdot)
\propto&
\;\alpha 
   \int 
   p(\mathbf{a}_{u} |\Muk, \Sk)  p(\Muk) p(\Sk)      
   \text{d}\Muk
   \text{d}\Sk\notag\\
   &\times
   \int 
   p(b_u | \muk, \sk)
   p(\muk)p(\sk) 
   \text{d}\muk
   \text{d}\sk \qquad \text{for } k=c+1
\end{align}
Unfortunately, these integrals are not analytically tractable because the product of the factors does not give a familiar distribution. \cite{Neal2000} proposes to create $m$ auxiliary empty components with parameters drawn from the base distribution, and then computing the likelihoods of $\mathbf{a}$ and $\mathbf{b}$ given those parameters. The higher the $m$, the closer we will be to the real integral and the less autocorrelated the cluster assignments will be. However, the equilibrium distribution of the Markov Chain is exactly correct for any value of $m$. To speed up the computations, $m$ is usually small. For our experiments, we chose $m=3$. That is, we generate $3$ additional empty tables each one with its own parameters  $\boldsymbol{\mu'}, \mathbf{S'}, \mu', s'$. We also run some experiments with $m=4$ and $m=5$, without observing significant differences neither in the clustering nor in the predictions, while it significantly increased the computational time. See \cite{Neal2000} for a more systematic study on the effect of $m$.

\subsection{Predictive distribution}
We are also interested in the ability of the model to predict new thread lengths. The posterior predictive distribution over the length of a new thread is:
\begin{align}
p(y_* | \mathbf{p_*}, & \mathbf{P}, \mathbf{y}) =
\int_{\boldsymbol{\theta}} 
p(y_* | \mathbf{p_*}, \boldsymbol{\theta})
p(\boldsymbol{\theta} | \mathbf{y}, \mathbf{P}) \text{d}\boldsymbol{\theta}
\end{align}
where $\mathbf{p_*}$ is the participation vector of the new thread, and $y_*$ its predicted length.
If we have samples $\boldsymbol{\theta}^{(1)},...,\boldsymbol{\theta}^{(N)}$ from the posterior $p(\boldsymbol{\theta}| \mathbf{y}, \mathbf{P})$, we can use them to approximate the predictive distribution:
\begin{align}
p(y_* | \mathbf{p_*}, & \mathbf{P}, \mathbf{y}) 
\approx
\frac{1}{N}
\sum_{i=1}^N
p(y_* | \mathbf{p_*}, \boldsymbol{\theta}^{(i)})
\label{eq:predictive_posterior_approx}
\end{align}
where $\boldsymbol{\theta}^{(i)}$ are the $i$-th samples of $\mathbf{b}$ and $\sigma_\text{y}$.


\section{Experiments}\label{sec:experiments}
We generated three scenarios to study in which situations dual-view models outperform single-view ones. The data reproduces the scenario of online forums presented in Section~\ref{sec:forums}. In the first scenario, users belong to five clusters and those who belong to the same cluster in one view also share the same cluster in the other view (\textit{agreement between views}). In the second scenario, users belong to five clusters in the behavior view but two of the clusters are completely overlapped in the feature view (\textit{disagreement between views}). In order to reproduce a more realistic clustering structure, in the last scenario user features and coefficients are taken from the \textit{iris dataset}. 

We will see in Section~\ref{sec:cost} that the main bottleneck of the model is the sampling of coefficients $b_1,...,b_U$ since they are obtained by sampling from a $U$-dimensional Gaussian distribution that requires, at each Gibbs iteration, inverting a $U\times U$ matrix to get its covariance. This issue would disappear if the inference of the behavioral function parameters for a user were independent from the parameters of the other users. In this paper, we use the \textit{iris} dataset to demonstrate the properties of the model as a whole, without making any statement on the convenience of the presented behavioral functions.

\subsection{Compared models}
We compared two dual-view models and one single-view model. We call them dual-fixed, dual-DP and single. The \texttt{dual-fixed} corresponds to the present model where the number of clusters is set to the ground truth (five clusters). The \texttt{dual-DP} corresponds to the present model where the number of clusters is also inferred (Section \ref{sec:CRP}). The \texttt{single} model corresponds to a Bayesian linear regression over the coefficients $\mathbf{b}$, which is equivalent to the behavior view specified in Section \ref{sec:forums_behaviors} where the number of clusters is set to one (that is, no clusters at all) and therefore there is no information flowing from the feature view to the behavior view; this model can only learn the latent coefficients $\mathbf{b}$.

Table \ref{tab:models} presents these three models as well as other related models that appear when blinding the models from one of the views. Note that we left out of the analysis those models that use clustering but are blinded of one view. The first two of these (IGMM and GMM), are regular clustering methods over feature vectors; we discarded them because they do not make inferences on latent behaviors. The last two (we call them latent-IGMM and latent-GMM) are Bayesian linear regressions where coefficients are assumed to come from a mixture of Gaussians; because these are in practice very similar to a simple Bayesian linear regression (they can be seen as Bayesian linear regressions with priors that tend to create groups of coefficients), we chose to benchmark only against the \texttt{single} model.

Posterior distributions of parameters are obtained by Gibbs sampling. We used the \texttt{coda} package \citep{coda, coda2015} in R \citep{R} to examine the traces of the Gibbs sampler. For the convergence diagnostics, we used Geweke's test available in the same package. After some initial runs and examinations of the chains to see how long it took to converge to the stationary distribution for the dual models, we observed that convergence for all the models is usually reached before 5,000 samples. Since we run a large number of automatized experiments, we set a conservative number of 30,000 samples for every run, from which the first 15,000 are discarded as burn-in samples. 
For the first two experiments we initialized our samplers with all users in the same cluster. For the \textit{iris} experiment we used the result of a k-means with 10 clusters over the feature view. We did not systematically benchmark the two initialisation strategies. Nevertheless, this second strategy is, in general, more convenient in order to arrive to the true target distribution within less iterations.
\begin{table}[ht]
\caption{Compared and related models. Both single models are the same since if the number of clusters is fixed to one they cannot use the feature view. The row marked as $-$ corresponds to a model that has no interest in this context since it simply finds the Gaussian distribution that best fits the observed features and makes neither clustering nor predictions.} 
 \begin{center}
   \tabcolsep = 1\tabcolsep
   \begin{tabular}{lccc}
   \hline\hline
                & features & behaviors & clusters\\
   \hline
   \textbf{dual-DP}    & yes & yes   & $\infty$\\
   \textbf{dual-fixed} & yes & yes   & fixed   \\
   \textbf{single}     & yes & yes   & 1 \\
   IGMM     & yes & no    & $\infty$ \\
   GMM        & yes & no    & fixed \\
   -          & yes & no    & 1 \\
   latent-IGMM & no  & yes   & $\infty$ \\
   latent-GMM    & no  & yes   & fixed \\
   \textbf{single}     & no  & yes   & 1 \\   
   \hline
   \end{tabular}
\label{tab:models}
 \end{center}
\end{table}

\subsection{Metrics}
We used two metrics for evaluation, one for clustering (within dual models) and one for predictions of thread lengths (within the three models). 

\paragraph{Metric for clustering:}
Clustering by mixtures models suffers from identifiability. The posterior distributions of $\mathbf{z}$ has $k!$ reflections corresponding to the $k!$ possible relabelling of the $k$ components. Due to this, different MCMC samples of $\mathbf{z}$ may come from different reflections making it hard to average the samples. A common practice is to summarize the pairwise posterior probability matrix of clustering, denoted by $\hat{\pi}$, that indicates the probability of every pair of users to be in the same component (no matter the label of the component). In order to obtain a full clustering $\mathbf{z}$ from $\hat{\pi}$, \cite{Dahl2006} proposes a \textit{least-squares model-based clustering} which consists of choosing as $\mathbf{z}$ the sample $\mathbf{z}^{(i)}$ whose corresponding pairwise matrix has the smaller least-squares distance to $\hat{\pi}$:

\begin{align}
\mathbf{z}_{LS} = \argmin_{\mathbf{z} \in \mathbf{z}^{(1)},..,\mathbf{z}^{(N)}} \sum_i^U \sum_j^U (\delta_{i,j}(\mathbf{z}) - \hat{\pi})^2
\end{align}
where $\delta_{i,j}(\mathbf{z})$ indicates whether $i$ and $j$ share the same component in $\mathbf{z}$.
Finally, to assess the quality of the proposed clustering $\mathbf{z}_{LS}$ we use the \textit{Adjusted Rand Index}, a measure of pair agreements between the true and the estimated clusterings.

\paragraph{Metric for predictions:}
For predictions, we use the \textit{negative loglikelihood}, 
which measures how likely the lengths are according to the predictive posterior:
\begin{align}
p(y^{\text{test}} | \mathbf{p}_t^{\text{test}}, \mathbf{P}^{\text{train}}, \mathbf{y}^{\text{train}})
\end{align}
and that can be approximated from Equation~\ref{eq:predictive_posterior_approx}. Negative loglikelihoods are computed on test sets of 100 threads.

\subsection{Agreement between views}
\begin{figure}
	\centering
	\includegraphics[width=1\textwidth]{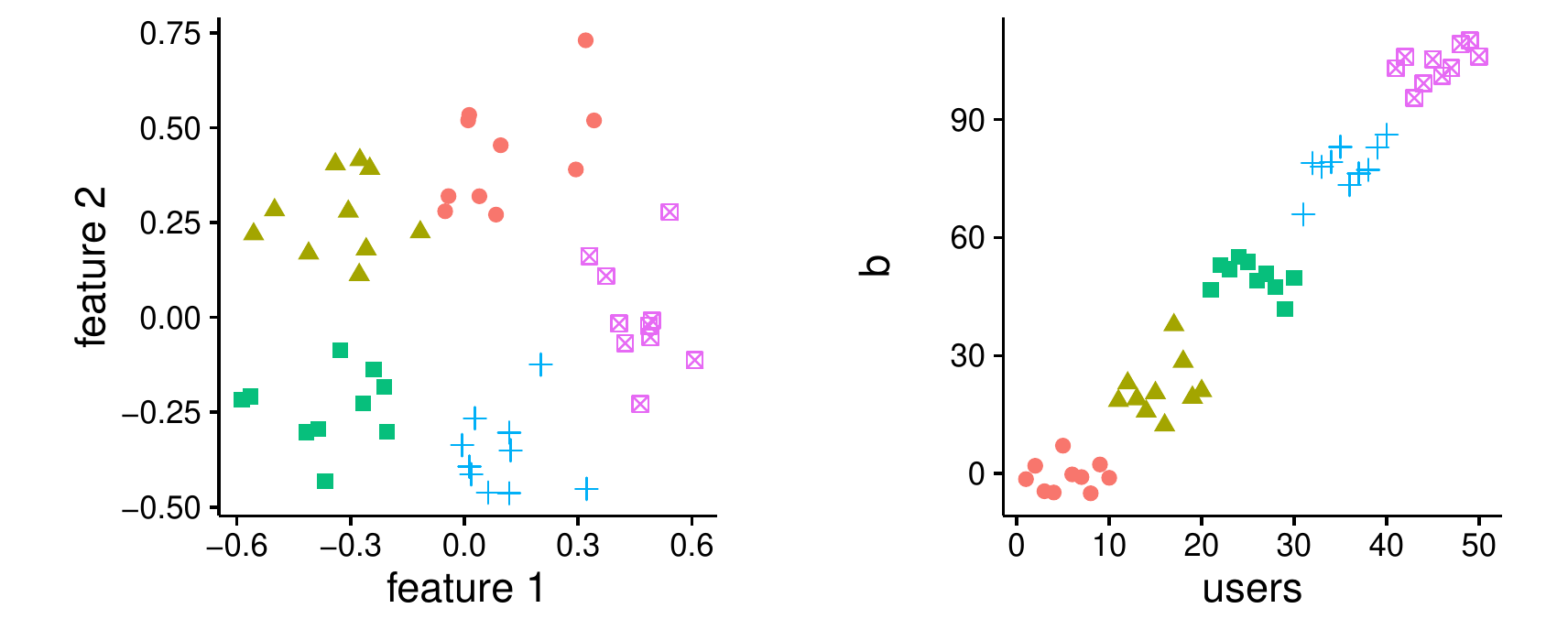}
	\caption{Dataset for agreement between the views. User features (left) and user coefficients (right). Every group (shape) of users has a well differentiated set of features and coefficients.}
	\label{fig:data_agreement}
\end{figure}

To set up the first scenario, for a set of $U$ users and five clusters we generated an assignment $z_u$ to one of the clusters so that the same number of users is assigned to every cluster.
Once all assignments $\mathbf{z}$ had been generated, we generated the data for each of the views. For the feature view, every user was given a two-dimensional feature vector $\mathbf{a}_u=(a_{u_1}, a_{u_2})^T$ drawn independently from:
	\begin{equation}
	\mathbf{a}_u \sim \mathcal{N}\left(\boldsymbol{\mu}_{z_u}, \Sigma_a \right)
	\end{equation}
where
	$\mu_{z_u}=\left(\cos(2\pi \frac{z_u}{5}), \sin(2\pi \frac{z_u}{5})\right)^T \text{for } z_u=1,..,5$
(see Figure~\ref{fig:data_agreement}). For the behavior view, every user was given a coefficient drawn independently from:
\begin{equation}
b_u \sim \mathcal{N}(-50 +25z_u, \sigma^2) 
\qquad\qquad \text{for } z_u=1,..,5
\end{equation}
where coefficients for users in the same cluster are generated from the same Normal distribution and the means of these distributions are equidistantly spread in a [-200,200] interval (see Figure~\ref{fig:data_agreement}).
To simulate users participations in a forum we generated, for each user $u$, a binary vector $\mathbf{p}_u = (p_{u1},...,p_{uT})^T$ of length $T$ that represents in which threads the user participated among the first $m$ posts. We supposed each user participated in average in half the threads.
\begin{equation}
p_{ut} \sim \text{Bernoulli}(0.5)
\end{equation}
Finally, we assumed that the final length of a thread is a linear combination of the coefficients of users who participated among the first posts:
	\begin{equation}
	y_t \sim \mathcal{N}(\mathbf{p}_t^T \mathbf{b}, \sigma_\text{y})
	\end{equation}

\begin{figure}
	\centering
	\includegraphics[width=1\textwidth]{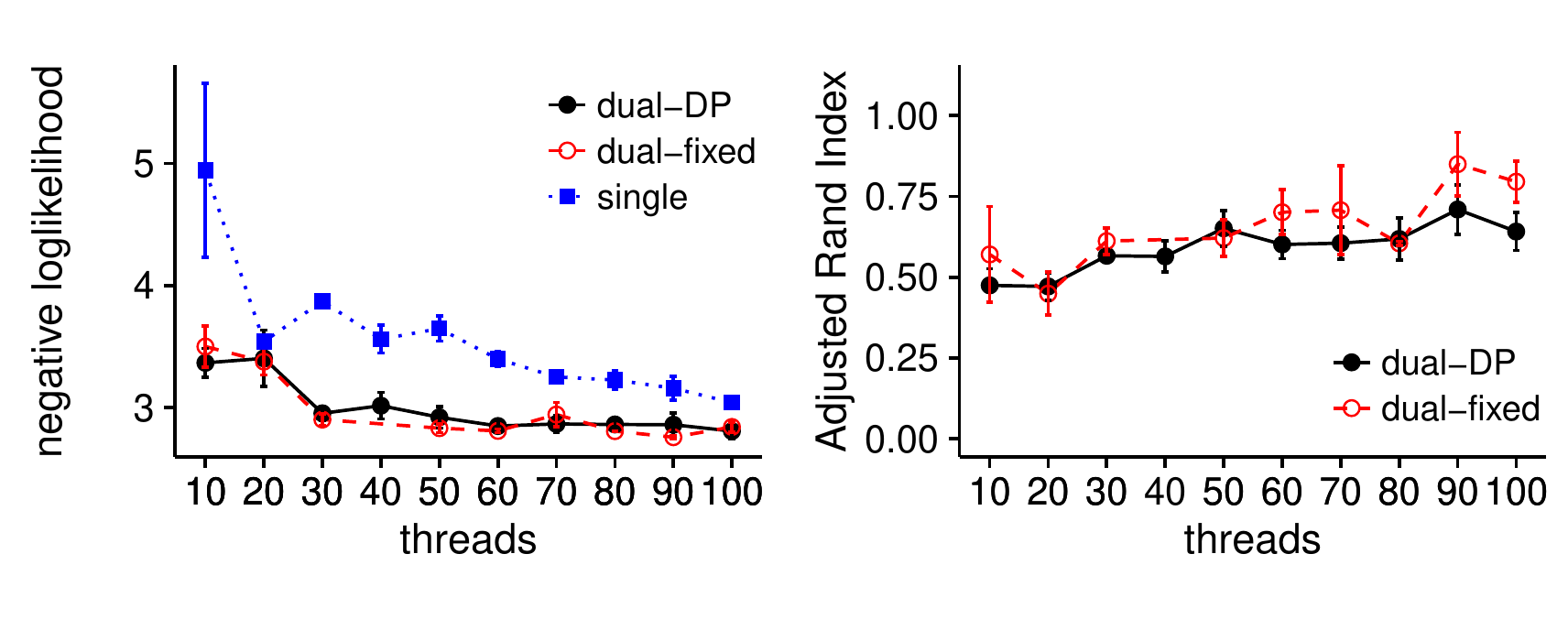}%
	\caption{Results for agreement between the views. Comparison of models under different threads/users ratios (50 users and variable number of threads). Means and standard errors over 5 runs.}
	\label{fig:results_agreement}
\end{figure}

If both views agree and there is enough data for both of them, we expect dual models to find the true clusters and true latent coefficients, and the single model to find also the true latent coefficients. In this case, the feature view brings no competitive advantage when there is enough information in the behavior view (and conversely, dual models should not outperform simple IGMM and GMM for the clustering task since there is enough information in the feature view).

On the contrary, when one of the views lacks information, then dual-view models should outperform single-view ones. In our model, the lack of information may come either from having too few threads to learn from or from having a high ratio of users versus threads since we have too many user behavior coefficients to learn. 

Figure \ref{fig:results_agreement} shows how increasing the threads vs users ratio affects the accuracy of each model. When the number of threads is too low with respect to the number of users neither view has enough information and thus the three models make bad predictions the inference is difficult for the three models. Yet, dual-view models need less threads than the single model to make good inferences. When the number of threads is high enough, the three models tend to converge.

The number of users and threads in the experiments ranges from 10 to 100 users and from 10 to 100 threads. We avoided larger numbers to prevent the experiments from taking too long. 30,000 iterations of the Gibbs sampler described in Section~\ref{sec:inference} for 50 users and 100 threads take around three hours in a Pentium Intel Core i7-4810MQ @2.80GHz. Nevertheless, the ratio users/threads remains realistic. In the real forums that we analyzed from \texttt{www.reddit.com} a common ratio is 1/10 for a window of one month.
\subsection{Disagreement between views}

\begin{figure*}
	\centering
	\includegraphics[width=1\textwidth]{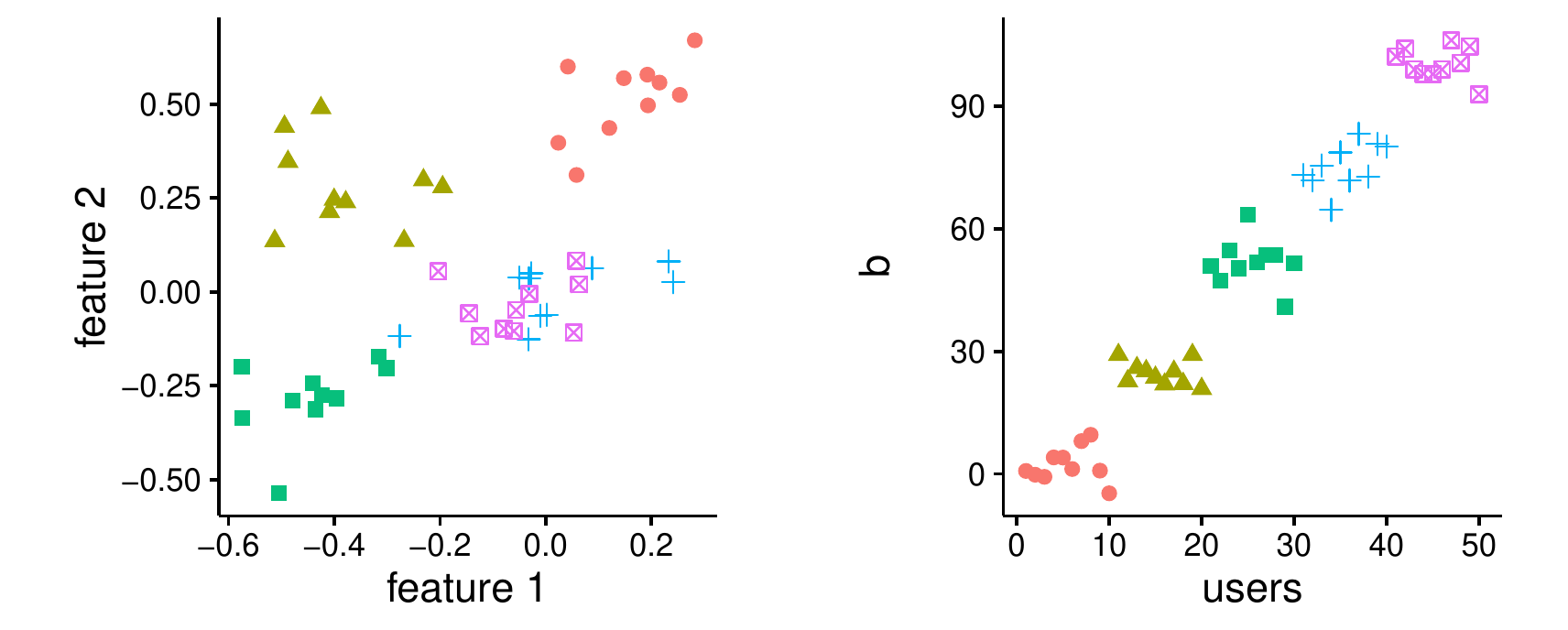}
	\caption{Dataset for disagreement between the views. User features (left) and user coefficients (right). Two of the groups (shapes) of users have similar features but different coefficients.}
	\label{fig:data_disagreement}
\end{figure*}

\begin{figure}
	\centering
	\subfloat[]{\includegraphics[width=0.33\textwidth]
		{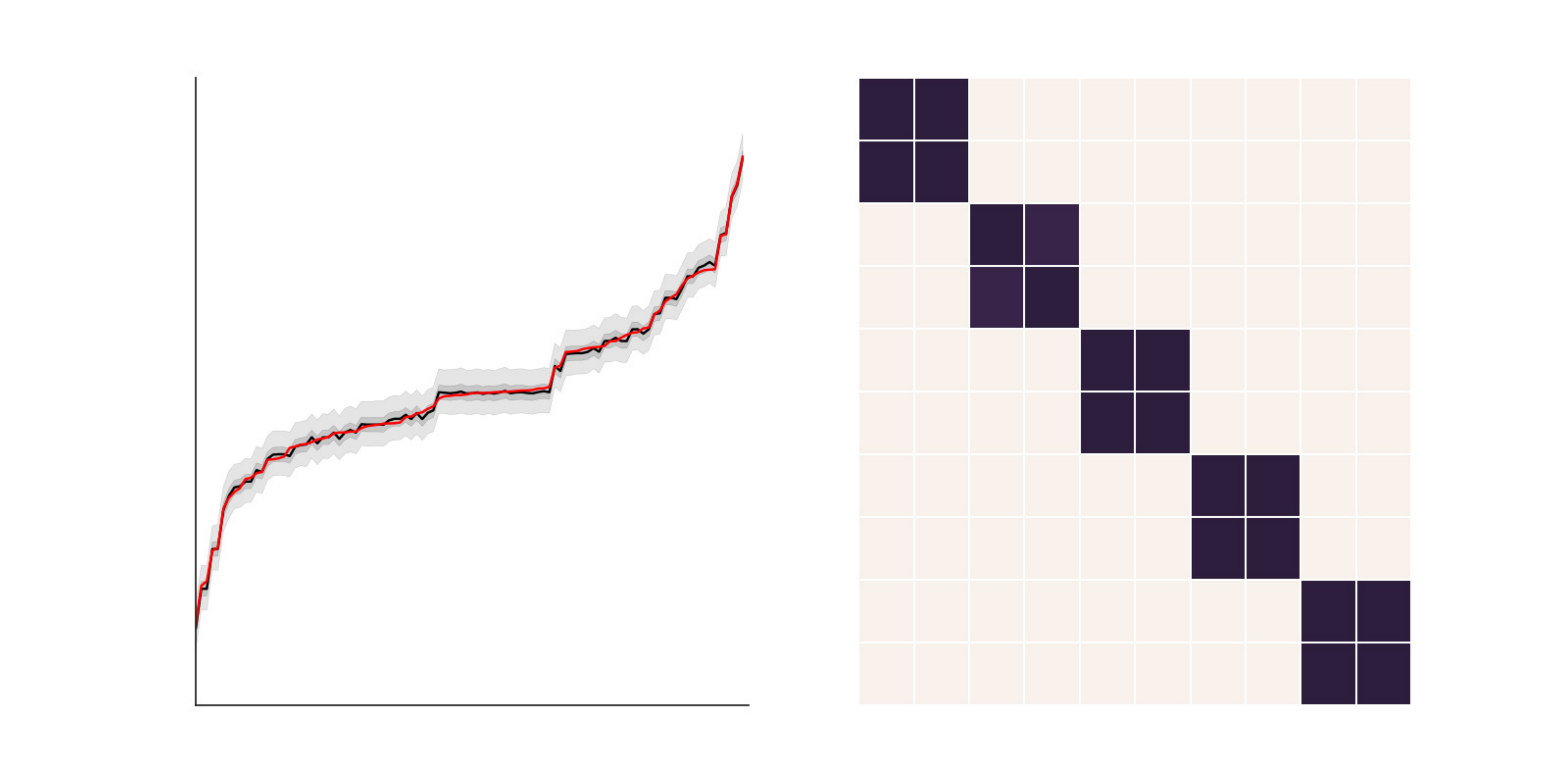}}%
	\subfloat[]{\includegraphics[width=0.33\textwidth]
		{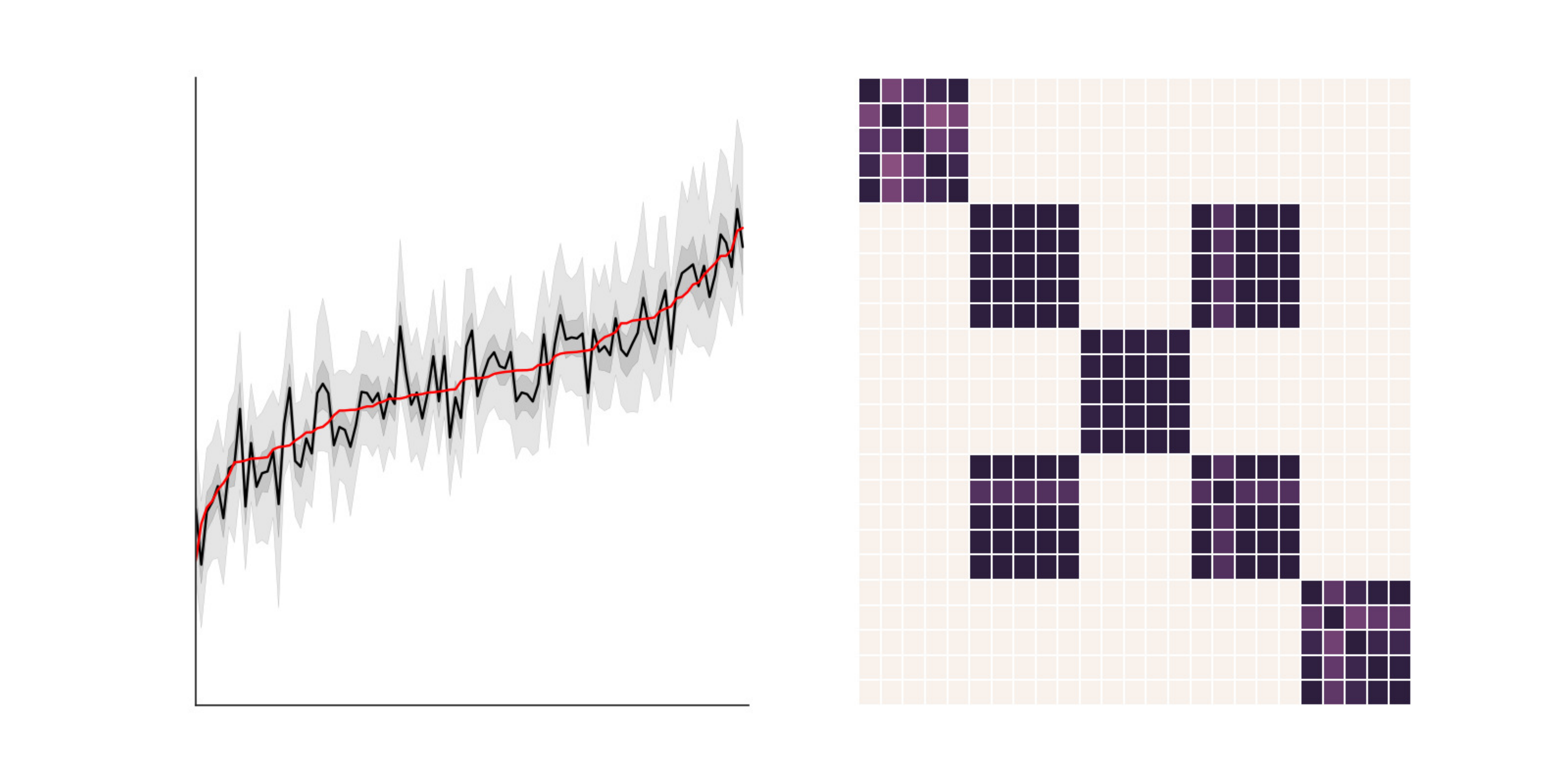}}%
	\subfloat[]{\includegraphics[width=0.33\textwidth]
		{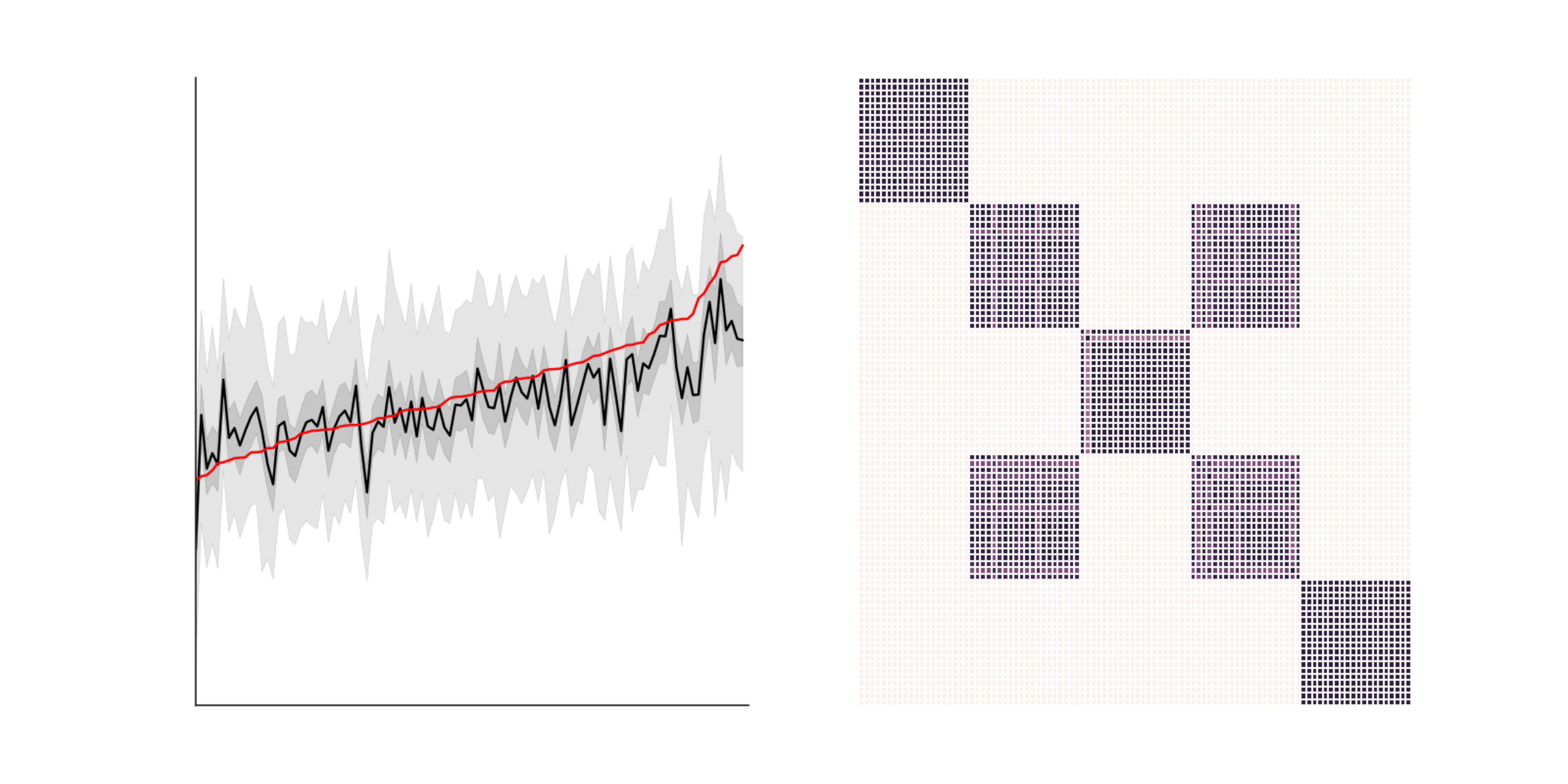}}
	\caption{Posteriors for DP-dual when the two views see a different number of clusters. (a) 10 users and 100 threads. (b) 25 users and 25 threads. (c) 100 users and 10 threads. Figures on the left: examples of posterior distributions of thread length predictions over 100 test threads with 50\% and 95\% credible intervals in test set with 50 users and 10, 50 and 100 threads. x-axis correspond to threads sorted by their (true) length while y-axis correspond to predicted thread lengths. True lengths are plotted in red (smoothest lines). Figures on the right: examples of posterior pairwise clustering matrices $\hat{\pi}$. x-axes and y-axes correspond to the users. A dark point means a high probability of being in the same cluster. The worst case is (c), which makes a similar clustering to (b) but worse predictions, because the feature view receives misguiding information from the behavior view and the number of threads is not high enough to compensate for it.}
	\label{fig:confused}
\end{figure}

\begin{figure*}
	\centering

	\includegraphics[width=1\textwidth]{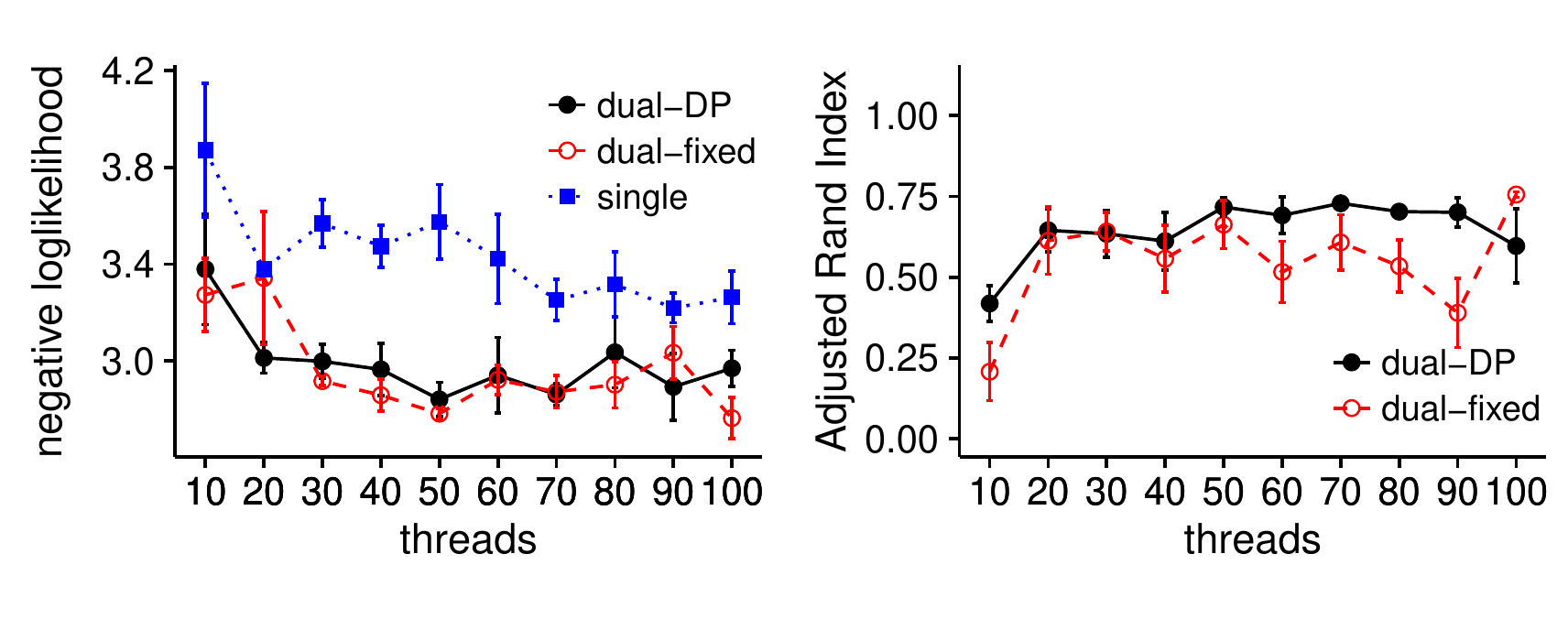}%
	\caption{Results for disagreement between the views. Comparison of models under different threads/users ratios (50 users and variable number of threads) when the two views see a different number of clusters. Means and standard errors over 5 runs.}
	\label{fig:results_disagreement}
\end{figure*}

If each view suggests a different clustering $\mathbf{z}$, dual models should find a consensus between them (recall Equation \ref{eq:z_posterior}). We generated a new dataset (Figure \ref{fig:data_disagreement}) where there are four clusters according to the feature view and five clusters according to the behavior view.

Figure~\ref{fig:confused} shows the posterior distributions (over thread lengths and over pairwise clustering) when (a) the behavior view has more information (b) both views lack data (c) the feature view has more information. By \textit{having more information} we mean that a view dominates the inference over the posterior of the clustering $\mathbf{z}$. 

\paragraph{(a) Lack of information in feature view:}
If the number of users is low but they participated in a sufficient number of threads, the behavior view (which sees five clusters) will be stronger than the feature view (which sees four clusters) and will impose a clustering in five groups. User coefficients (used for thread length predictions) are also well estimated since the number of threads is high enough to learn them despite the lack of help from the other view (Figure~\ref{fig:confused}a).

\paragraph{(b) Lack of information in both views:}
In the middle ground where neither view has enough evidence, the model recovers four clusters and the predictive posterior over thread lengths gets wider though still making good inferences (Figure~\ref{fig:confused}b).

\paragraph{(c) Lack of information in behavior view:}
If the number of users is high and the number of threads is low, the feature view (four clusters) will have more influence in the posterior than the behavior view (five clusters), (Figure~\ref{fig:confused}c). Predictions get worse because the model imposes a four clusters prior over coefficients that are clearly grouped in five clusters.

In order to compare between the performance in case of agreement and the performance in case of disagreement, we repeated the experiments of the last section with the current configuration. 
Figure~\ref{fig:results_disagreement} shows the performance of the models for 50 users and a different number of threads. While predictions and clustering improve with the number of threads, clustering with a small number of threads is worse in case of disagreement since the feature view imposes its 4 clusters vision. To recover the five clusters we would need either more  threads or less users.

For the predictions, the dual models still outperform the single one because the feature view mostly agrees with the behavior view except for the users in one of the clusters. If all user features were in the same cluster, (no clustering structure) the performance of the predictions would be similar for the three models since the feature view would add no extra information. If we randomize the features so that, for instance, there are five clusters in the feature view that are very different from the clusters in the behavior view, we may expect the dual-view  models to give worse predictions than the single-view one in those cases where they now perform better. In those cases, dual-models would be getting noise in the feature view (or very bad priors) and only a large enough number of threads could compensate for it.

\subsection{Iris dataset}
\begin{figure}
	\centering
	\includegraphics[width=0.48\textwidth]{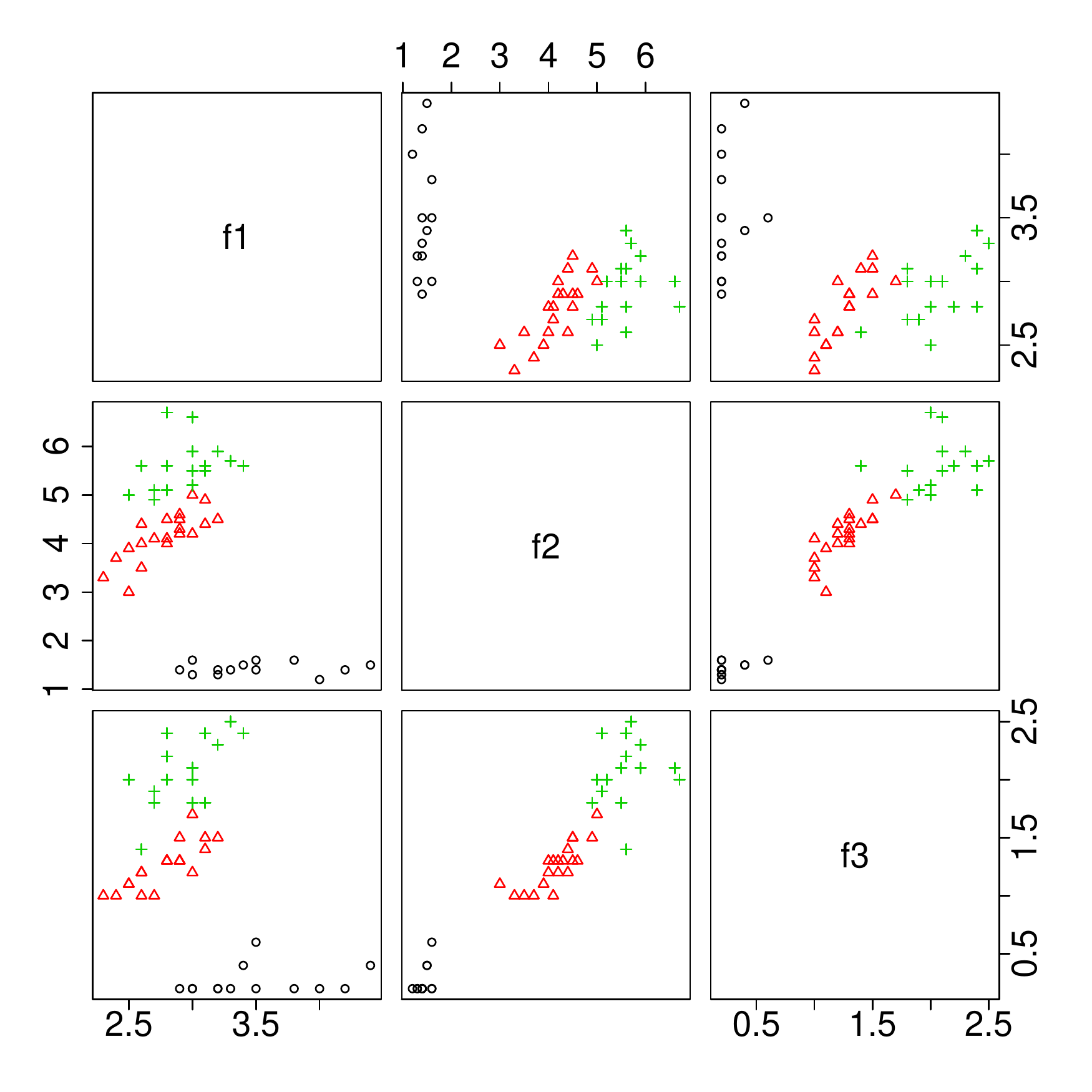}%
	\includegraphics[width=0.5\textwidth]{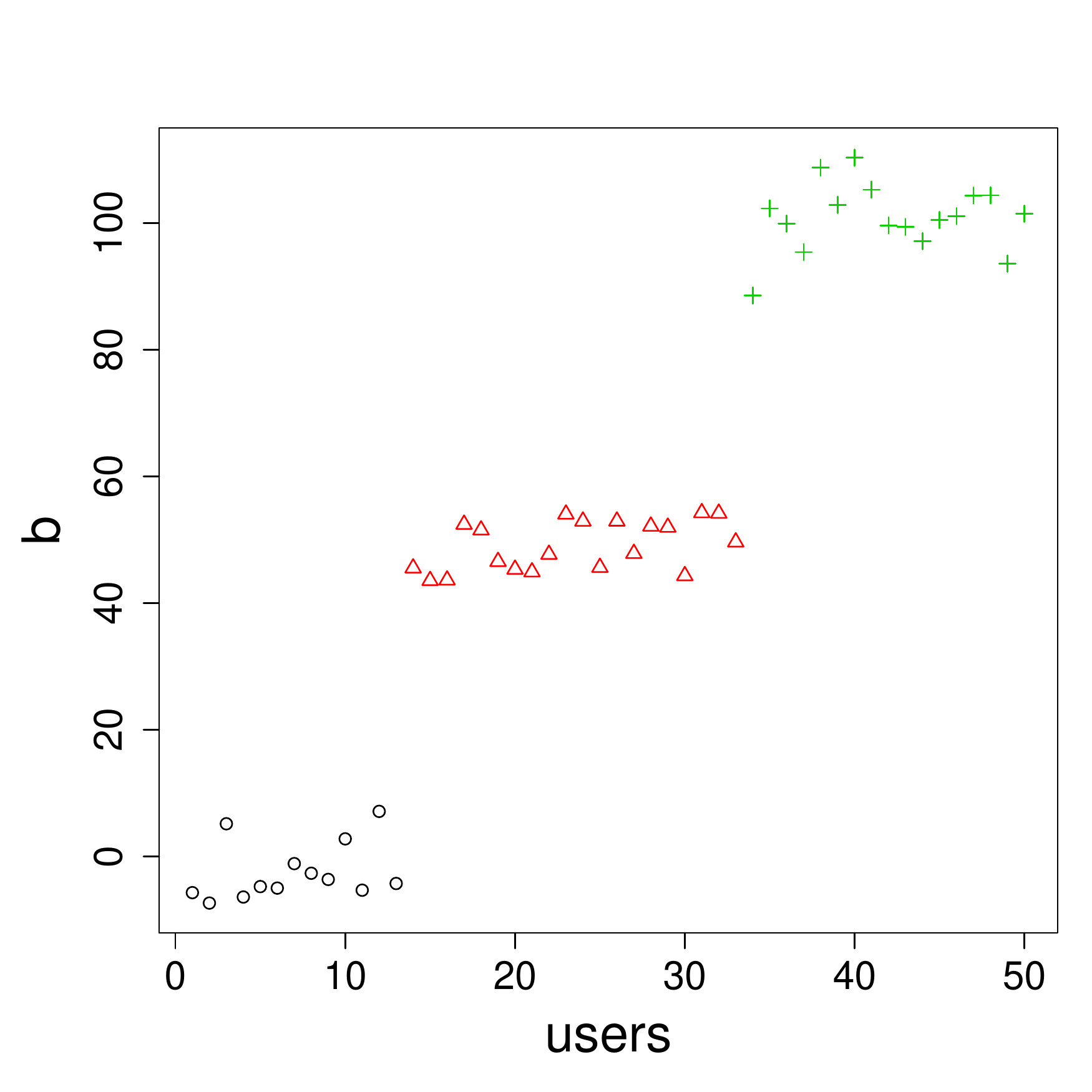}
	\caption{Iris dataset. User features (left) and synthetic user coefficients (right)}
	\label{fig:iris_data}
\end{figure}
\begin{figure}
	\centering
	\includegraphics[width=1\textwidth]{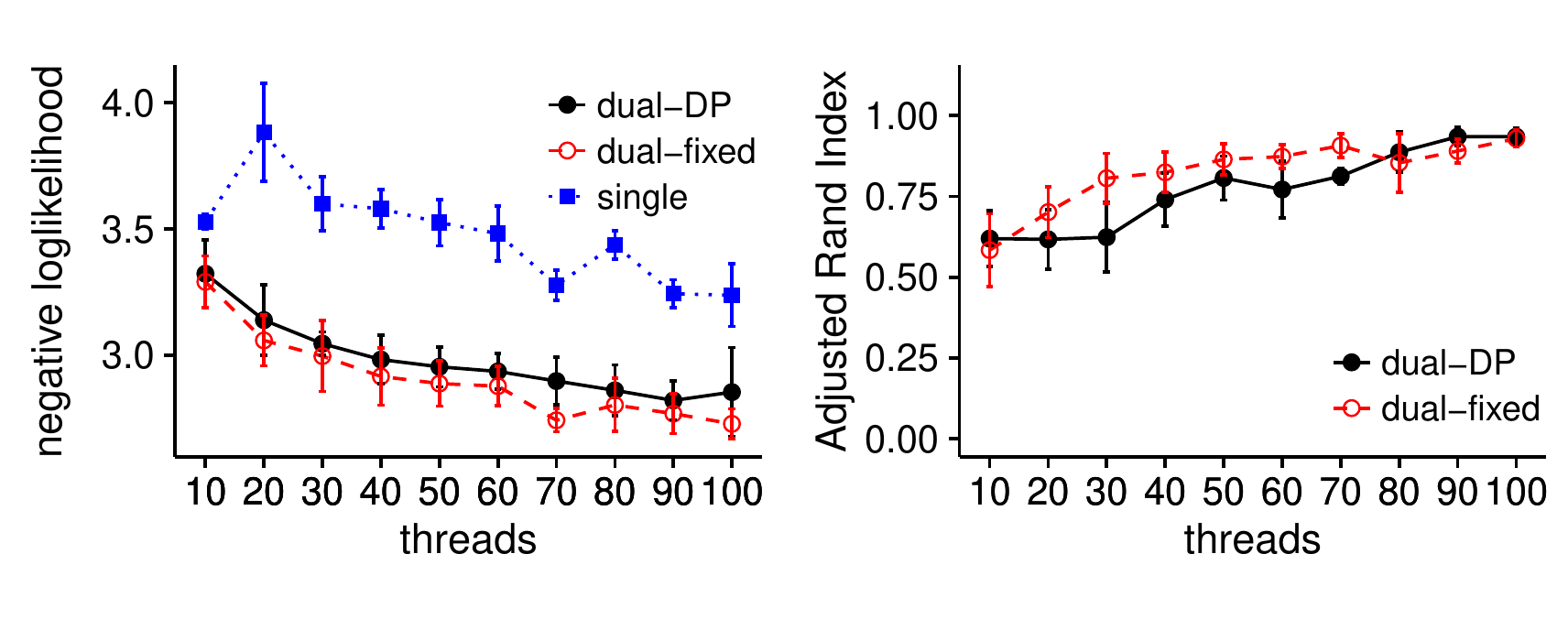}
	\caption{Results for the iris dataset. Comparison of models under different threads/users ratios (50 users and variable number of threads). Means and standard errors over 5 runs.}
	\label{fig:iris_results}
\end{figure}

\begin{figure}
	\centering
	\includegraphics[width=0.5\textwidth]{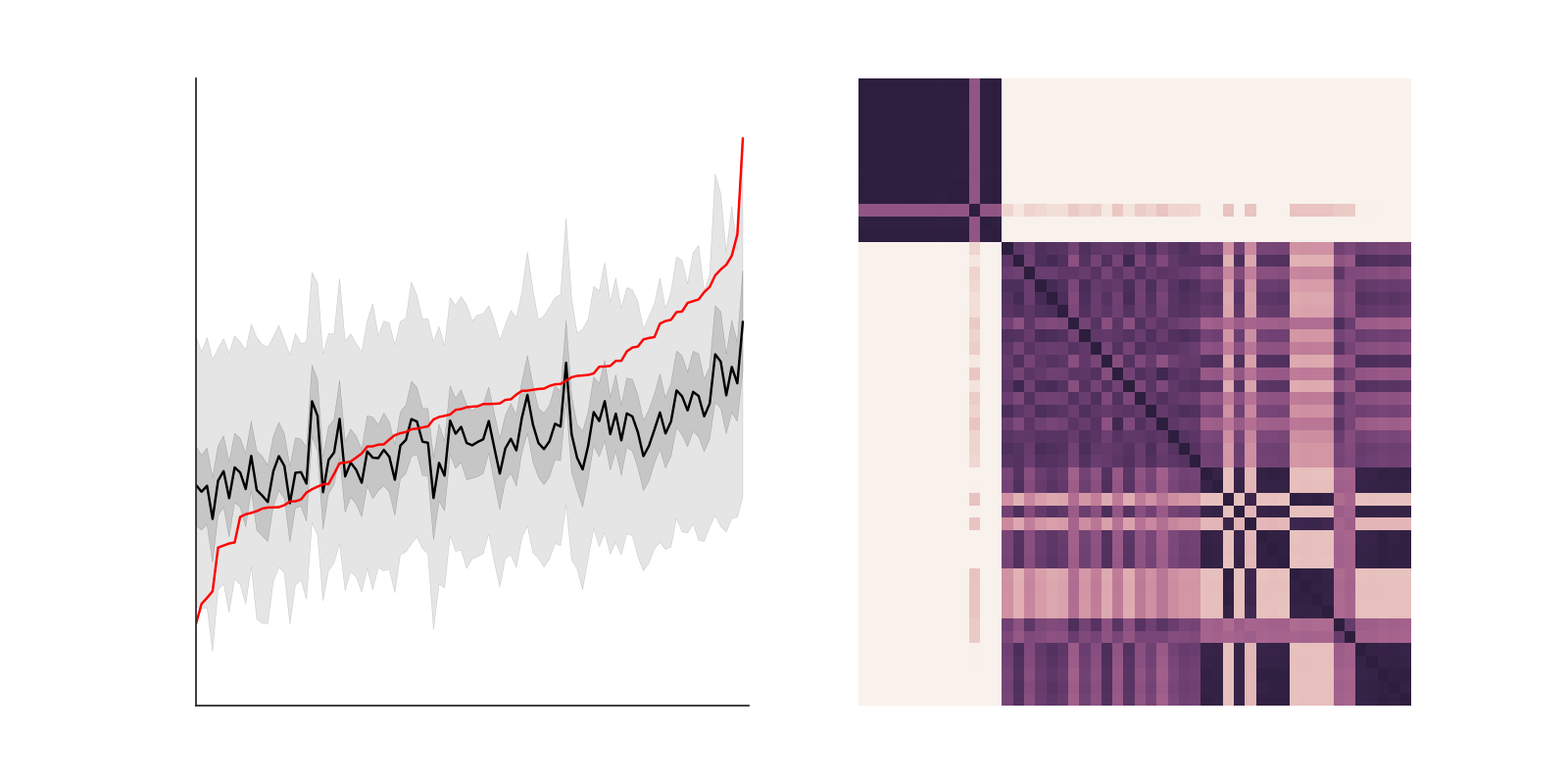}%
	\includegraphics[width=0.5\textwidth]{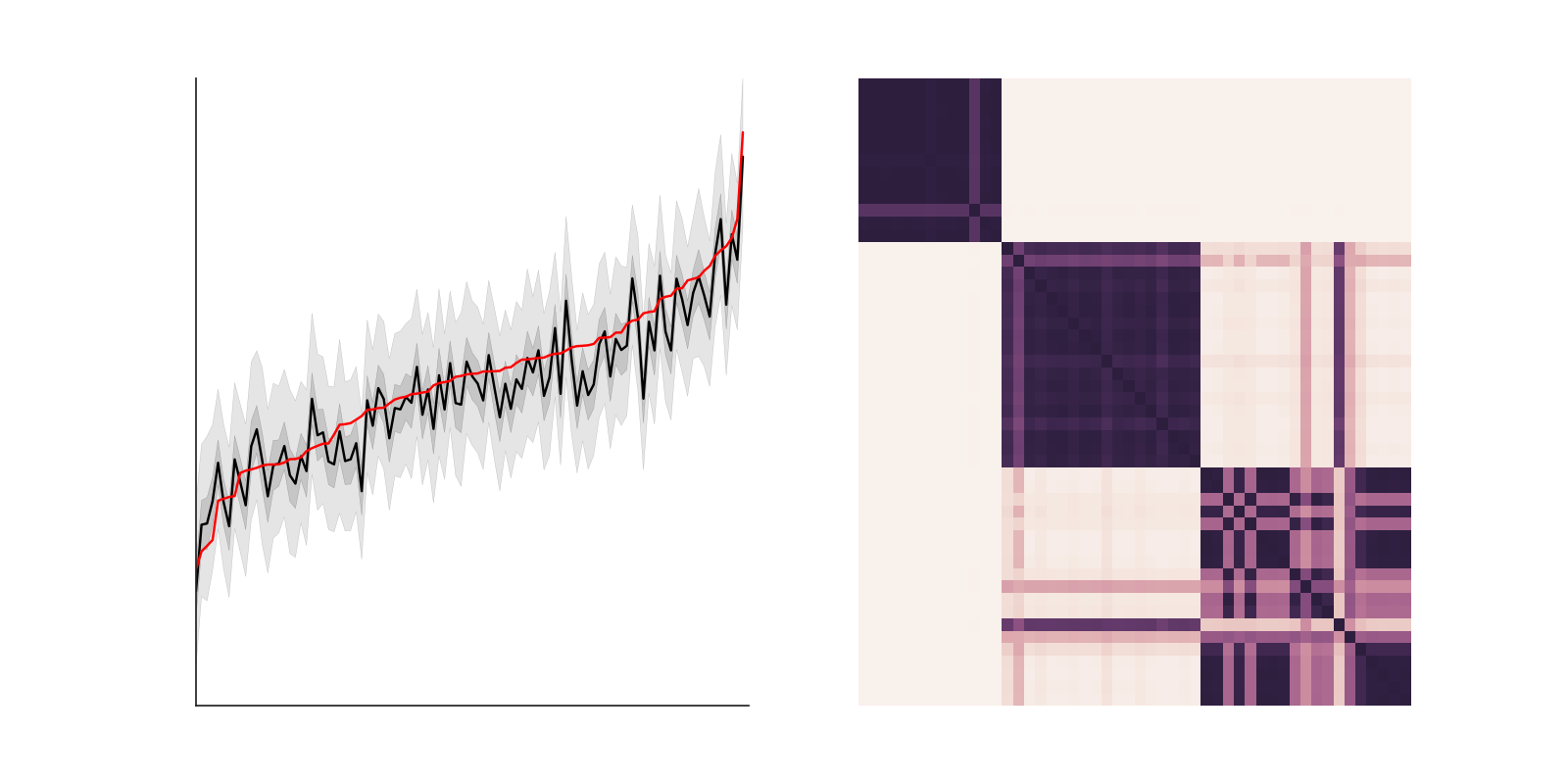}
	\caption{Predictive posteriors given by the \texttt{DP-dual} model over thread lengths and pairwise clustering matrices with 50 users and a training set of 10 threads (left) and 100 threads (right). Shadowed areas indicate 95\% and 50\% credible intervals. True lengths are plotted in red (smoothest lines). As the number of thread increases, the model starts to see the three cluster structure as well as to make better predictions.}
	\label{fig:iris_posteriors}
\end{figure}

To reproduce a more realistic clustering structure we performed a last experiment based on the \textit{iris} dataset. We used the \textit{iris} data available in R, which corresponds to the original dataset reported in \cite{IrisData1935}. In our experiment, features correspond to three of the features of the \textit{iris} dataset (Figure \ref{fig:iris_data}). We chose three out of the four features (sepal width, petal length and petal width) as well as a random subset of 50 observations so that the clustering task is harder if we only use the feature view. The coefficients of the behavior view are similar to those used in the former experiments. The selected observations are assigned to the same cluster than in the \textit{iris} dataset (species). We run a standard EM-GMM, from the R package \texttt{mclust} \citep{mclust}, over the features to have an idea of what we should expect from our model when there are almost no threads and the coefficients are difficult to infer. We also run the same EM-GMM over the features and the true coefficients to have an idea of what we should expect from our model when the number of threads is high enough to make a good inference of the coefficients. This gave us an ARI of 0.48 and 0.79, respectively. Indeed, starting nearer to 0.48 when the number of threads is small, our model gets closer to 0.79 as we keep adding threads (Figure \ref{fig:iris_results}). Of course, the inference of the coefficients and thus the predictions over the test set also improve by increasing the number of threads. Since the single model does not take profit of the feature view, it needs more threads to reach the same levels than its dual counterparts. Figure~\ref{fig:iris_posteriors} shows two examples of confusion matrices and predicted lengths for 10 and 100 threads.
\begin{figure}
	\centering
	\includegraphics[width=1\textwidth]{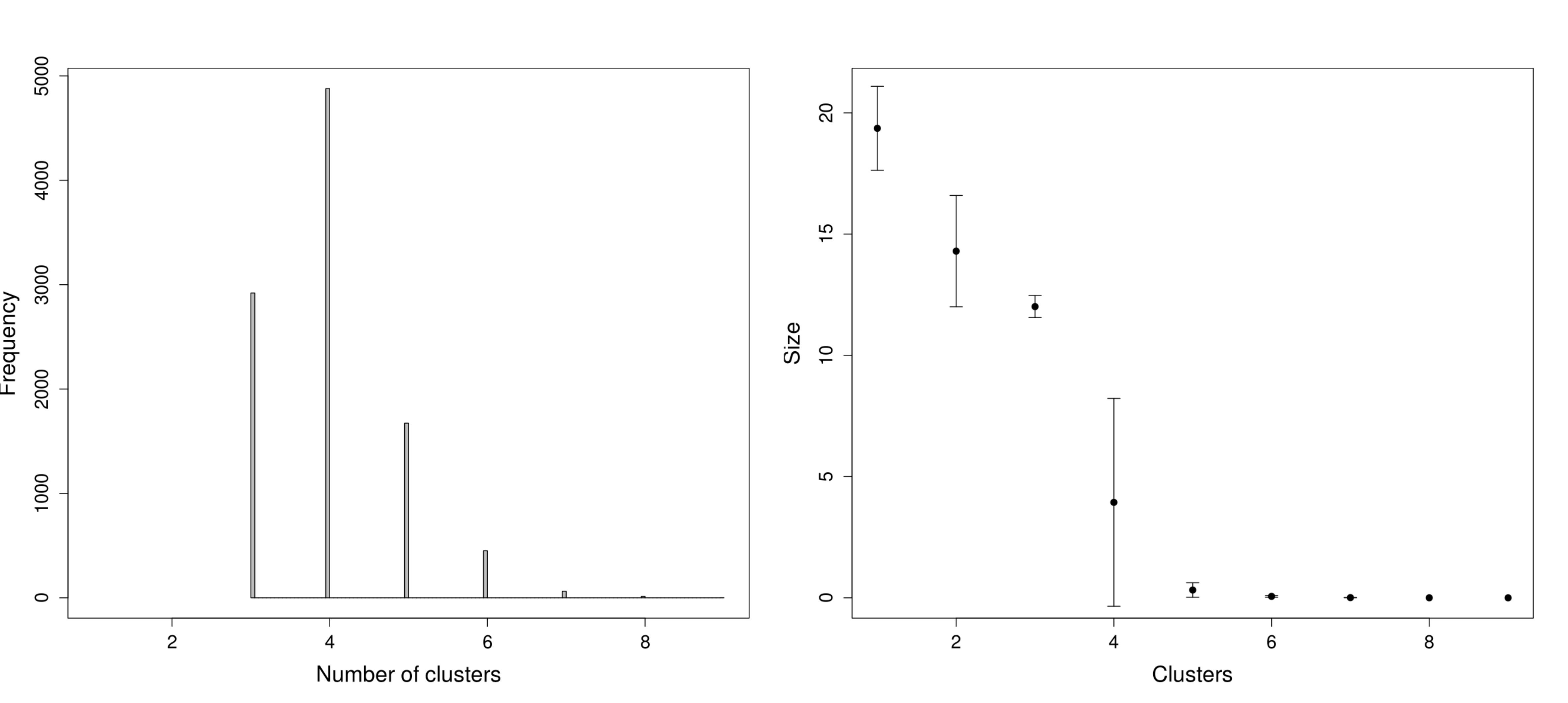}
	\caption{Left: histogram of the number of active clusters estimated by the \texttt{DP-model} in the \textit{iris} scenario with 50 threads. Right: mean and standard deviations of the number of users assigned to each cluster during the chain. Most users are assigned to the three major clusters and a small group of users is assigned to a fourth cluster.}
	\label{fig:nclusters}
\end{figure}

Figure~\ref{fig:nclusters} shows the histogram of the number of clusters within the MCMC chain and the distribution of cluster sizes. The model infers three clusters but it also places some probability over a two clusters structure due to the closeness of two of the clusters in the feature view.  

\subsection{Computational cost}\label{sec:cost}
We analyzed the computational cost of the \texttt{dual-DP} model since it is the most complex of the three compared. Unlike the \texttt{single} model, it makes inferences in the two views, meaning about twice the number of variables.  And unlike the \texttt{fixed-dual}, it has to infer the number of cluster and does it by creating $m$ empty candidate clusters every time we sample a cluster assignment for a user at each iteration of the Gibbs sampler. This means creating $U \times iterations \times m$ empty clusters and computing, as many times, whether a user belongs to one of these auxiliary clusters ($m$ possible extra clusters at each iteration), which makes it the slowest of the three models in terms of time per iteration.

We look at the autocorrelation time to estimate the distance between two uncorrelated samples:
\begin{equation*}
	\tau = 1 + 2\sum_{n=1}^{1000} |\rho_n|
\end{equation*} 
where $\rho_n$ is the autocorrelation at lag $n$. The variable with the higher autocorrelation is $\muo$ which has an autocorrelation time of 79. Since we drop the first 15000 samples for burn-in, we get an Effective Sample Size of 189 independent samples.

The bottlenecks of the algorithm are computing the likelihoods of the features and the coefficients given a cluster assignment, and the sampling of the coefficients. The sampling of the coefficients is relatively slow because it implies sampling from a multivariate Gaussian distribution with a $U\times U$ covariance matrix (see Appendix). This is due to the fact that the coefficient of each user depends on the coefficients of the users who have co-participated in the same thread. Note that this bottleneck would disappear in other scenarios where the inference of the behavioral function of a user is independent from the other users.

\subsection{Summary of the experiments}
To summarize, the experiments show on the one hand that dual-view models outperform single-view ones when users can be grouped in clusters that share similar features and latent behavioral functions and, on the other hand, that even when this assumption is not true, as long as there is enough data, the inference will lead to a consensual partition and a good estimation of latent functions. Indeed, each view acts as a prior, or a regularization factor, on the other view. Good priors improve inference, while bad priors misguide the inferences unless there is sufficient amount of evidence to ignore the prior.

\section{Conclusions}\label{sec:conclusions}

We presented a dual-view mixture model to cluster users based on features and latent behavioral functions. Every component of the mixture model represents a probability density over two \textit{views}: a feature view for observed user attributes and a
behavior view for latent behavioral functions that are indirectly observed through user actions or behaviors. The posterior distribution of the clustering represents a consensual clustering between the two views. Inference of the parameters in each view depends on the other view through this common clustering, which can be seen as a proxy that passes information between the views. An appealing property of the model is that inference on latent behavioral functions may be used to make predictions of users future behaviors. We presented two versions of the model: a parametric one where the number of clusters is treated as a fixed parameter and a nonparametric, based on a Dirichlet Process, where the number of clusters is also inferred.

We have adapted the model to a hypothetical case of online forums where behaviors correspond to the ability of users to generate long discussions. We clustered users and inferred their behavioral functions in three datasets to understand the properties of the model. We inferred the posteriors of interest by Gibbs sampling for all the variables but two of them which were inferred by Adapted Rejection Sampling. Experiments confirm that the proposed dual-view model is able to learn with less instances than its single-view counterpart due to the fact that dual-view models use more information. Moreover, inferences with the dual-view model based on a Dirichlet Process are as good as inferences with the parametric model even if the latter knows the true number of clusters.

In our future research we plan to adapt and apply the model to more realistic tasks such as learning users preferences based on choices and users features. Particularly, we would like to compare our model to that of \cite{Abbasnejad2013a} in the \textit{sushi} dataset \citep{Kamishima2009}. Also, it
might be interesting to consider latent functions at a group level, that is, that users in the same cluster share \textit{exactly} the same latent behavior. Not only it would reduce the computational cost but, if we have few data about every user, a group-level inference may also be more grounded and statistically sound.

Finally, in order to apply the model to large scale data we will also explore alternative and faster inference methods such as Bayesian Variational Inference.

\newpage
\appendix
\noindent \textbf{\Large Appendix}
\vspace{-0.1cm}
\section{Chinese Restaurant Process}
In this section we recall the derivation of a Chinese Restaurant Process. Such a process will be used as the prior over cluster assignments in the model. This prior will then be updated through the likelihoods of the observations through the different views. 

Imagine that every user $u$ belongs to one of $K$ clusters. $z_u$ is the cluster of user $u$ and $\mathbf{z}$ is a vector that indicates the cluster of every user. Let us assume that $z_u$ is a random variable drawn from a multimomial distribution with probabilities $\boldsymbol{\pi}= (\pi_1,...,\pi_K)$. Let us also assume that the vector $\boldsymbol{\pi}$ is a random variable drawn from a Dirichlet distribution with a symmetric concentration parameter $\boldsymbol{\alpha} = (\alpha/K,...,\alpha/K)$. We have: 
\begin{align*}
z_u | \boldsymbol{\pi} &\sim \text{Multinomial}(\boldsymbol{\pi})\notag\\
\boldsymbol{\pi} &\sim \text{Dirichlet}(\boldsymbol{\alpha})
\end{align*}
The marginal probability of the set of cluster assignments $\mathbf{z}$ is:
\begin{align*}
p(\mathbf{z}) =& 
\int \prod_{u=1}^U p(z_u | \boldsymbol{\pi})p(\boldsymbol{\pi} | \boldsymbol{\alpha})
\text{d}\boldsymbol{\pi}\\
=&\int 
   \prod_{i=1}^K \pi_i^{n_i} 
   \frac{1}{B(\boldsymbol{\alpha})}
   \prod_{j=1}^K \pi_j^{\alpha/K-1}
   \text{d}\boldsymbol{\pi}\\
=&
\frac{1}{B(\boldsymbol{\alpha})}
\int 
   \prod_{i=1}^K \pi_i^{\alpha/K + n_i - 1}
   \text{d}\boldsymbol{\pi}
\end{align*}
where $n_i$ is the number of users in cluster $i$ and $B$ denotes the Beta function. Noticing that the integrated factor is a Dirichlet distribution with concentration parameter $\boldsymbol{\alpha} + \mathbf{n}$ but without its normalizing factor:
\begin{align*}
p(\mathbf{z})=&
\frac{
B(\boldsymbol{\alpha} + \mathbf{n})
}{
B(\boldsymbol{\alpha})
}
\int 
   \frac{1}{
   B(\boldsymbol{\alpha + \mathbf{n}})
   }
   \prod_{i=1}^K \pi_i^{\alpha/K + n_i - 1}
   \text{d}\boldsymbol{\pi}\\
=& 
\frac{
B(\boldsymbol{\alpha} + \mathbf{n})
}{
B(\boldsymbol{\alpha})
}
\end{align*}
which expanding the definition of the Beta function becomes:
\begin{align}
p(\mathbf{z})=
\frac{
\prod_{i=1}^K \Gamma(\alpha/K + n_i)
}
{
\Gamma \left(\sum_{i=1}^K \alpha/K + n_i \right)
}
\frac{
\Gamma \left(\sum_{i=1}^K \alpha/K \right)
}
{
\prod_{i=1}^K \Gamma(\alpha/K)
}
= 
\frac{
	\prod_{i=1}^K \Gamma(\alpha/K + n_i)
}
{
	\Gamma \left(\alpha + U \right)
}
\frac{
	\Gamma \left(\alpha \right)
}
{
	\prod_{i=1}^K \Gamma(\alpha/K)
}
\label{eq:p_z}
\end{align}
where $U=\sum_{i=1}^{K}n_i$. Note that marginalizing out $\boldsymbol{\pi}$ we introduce dependencies between the individual clusters assignments under the form of the counts  $n_i$. The conditional distribution of an individual assignment given the others is:
\begin{align}
\label{eq:z_cond}
p(z_u  = j| \mathbf{z_{-u}}) 
=
\frac{p(\mathbf{z})}
{p(\mathbf{z}_{-u})}
\end{align}
To compute the denominator we assume cluster assignments are exchangeable, that is, the joint distribution $p(\mathbf{z})$ is the same regardless the order in which clusters are assigned. This allows us to assume that $z_u$ is the last assignment, therefore obtaining $p(\mathbf{z_{-u}})$ by considering how Equation $\ref{eq:p_z}$ before $z_u$ was assigned to cluster $j$. 
\begin{align}
\label{eq:p_z_minus}
p(\mathbf{z}_{-u}) =& 
\frac
{
\Gamma(\alpha/K + n_j-1)
\prod_{i\neq j} \Gamma(\alpha/K + n_i)
}
{\Gamma
\left(
\alpha + U -1 
\right)
}
\frac{
\Gamma \left(\alpha \right)
}
{
\prod_{i=1} \Gamma(\alpha/K)
}
\end{align}
And finally plugging Equations \ref{eq:p_z_minus} and \ref{eq:p_z} into Equation \ref{eq:z_cond}, and cancelling out the factors that do not depend on  the cluster assignment $z_u$, and finally using the identity $a \Gamma(a) = \Gamma(a+1)$ we get:

\begin{align*}
p(z_u = j| \mathbf{z}_{-u}) 
&=
\frac
{\alpha/K + n_j-1}
{\alpha + U -1}
=
\frac
{\alpha/K + n_{-j}}
{\alpha + U -1} 
\end{align*}

where $n_{-j}$ is the number of users in cluster $j$ before the assignment of $z_u$.

The Chinese Restaurant Process is the consequence of considering $K \rightarrow \infty$. For clusters where $n_{-j}>0$, we have:
\begin{align*}
p(z_u = j \text{ s.t } n_{-j}>0 | \mathbf{z}_{-u}) 
&=
\frac
{n_{-j}}
{\alpha + U -1}
\end{align*}
and the probability of assigning $z_u$ to any of the (infinite) empty clusters is: 
\begin{align*}
p(z_u = j \text{ s.t } n_{-j}=0 | \mathbf{z_{-u}}) 
=\;& \lim_{K\rightarrow \infty}
(K - p)\frac
{\alpha/K}
{\alpha + U -1} 
= 
\frac
{\alpha}
{\alpha + U -1} 
\end{align*}
where $p$ is the number of non-empty components.
It can be shown that the generative process composed of a Chinese Restaurant Process were every component $j$ is associated to a probability distribution with parameters $\boldsymbol{\theta}_j$ is equivalent to a Dirichlet Process.

\section{Conditionals for the feature view}
In this appendix we provide the conditional distributions for the feature view to be plugged into the Gibbs sampler. Note that, except for $\betaoa$, conjugacy can be exploited in every case and therefore their derivations are straightforward and well known. The derivation for $\betaoa$ is left for another section:
\subsection{Component parameters}
\subsubsection*{Components means $p(\Muk | \cdot )$:}
\begin{align*}
p(\Muk | \cdot ) 
&\propto
p\left(\Muk | \Muo, \invRo\right) 
\prod_{u \in k} p\left(\mathbf{a}_u | \Muk, \Sk, \mathbf{z}\right)\\
&\propto
\mathcal{N}\left(\Muk | \Muo, \invRo\right) 
\prod_{u \in k} \mathcal{N}\left(\mathbf{a}_u | \Muk, \Sk\right)\\
&=
\mathcal{N}(\boldsymbol{\mu', \Lambda'})
\end{align*}
where:
\begin{align*}
\boldsymbol{\Lambda'} &= \Ro + n_k \Sk\\ 
\boldsymbol{\mu'} &= \boldsymbol{\Lambda'^{-1}} \left(\Ro \Muo + \Sk \sum_{u\in k} \mathbf{a}_u\right)
\end{align*}

\subsubsection*{Components precisions $p(\Sk | \cdot )$:}
\begin{align*}
p(\Sk | \cdot ) 
\propto\;& 
p\left(\Sk |\betaoa, \Wo\right)
\prod_{u \in k} p\left(\mathbf{a}_u | \Muk, \Sk, \mathbf{z}\right)\\
\propto\;&
\mathcal{W}\left(\Sk |\betaoa, (\betaoa
\Wo)^{-1}\right)
\prod_{u \in k} \mathcal{N}\left(\mathbf{a}_u | \Muk, \Sk\right)\\
=\;& \mathcal{W}(\beta', \mathbf{W}')
\end{align*}
where:
\begin{align*}
\beta' &= \betaoa + n_k\\
\mathbf{W}' &= 
\left[ \betaoa\Wo + \sum_{u \in k} (\mathbf{a}_u - \Muk)(\mathbf{a}_u- \Muk)^T  \right]^{-1}
\end{align*}

\subsection{Shared hyper-parameters}
\subsubsection*{Shared base means $p(\Muo | \cdot)$:}
\begin{align*}
p(\Muo | \cdot) 
&\propto
p\left(\Muo | \boldsymbol{\mu_a}, \boldsymbol{\Sigma_a}\right)
\prod_{k = 1}^K p\left(\Muk | \Muo, \Ro \right)  \\
&\propto
\mathcal{N}\left(\Muo | \boldsymbol{\mu_a, \Sigma_a}\right)
\prod_{k = 1}^K\mathcal{N}\left(\Muk | \Muo, \invRo\right)  \\
&=\mathcal{N}\left(\boldsymbol{\mu'}, \boldsymbol{\Lambda'}^{-1}\right)
\end{align*}
where:
\begin{align*}
\boldsymbol{\Lambda'} &= \boldsymbol{\Lambda_{a}} + K \Ro\\ 
\boldsymbol{\mu'} &= \boldsymbol{\Lambda'}^{-1} \left(\boldsymbol{\Lambda_{a}} \boldsymbol{\mu_{a}} + K \Ro \overline{\Muk}\right)
\end{align*}

\subsubsection*{Shared base precisions $p(\Ro | \cdot)$:}
\begin{align*}
p(\Ro | \cdot) 
\propto\;&
p\left(\Ro | D, \boldsymbol{\Sigma_a^{-1}}\right) 
\prod_{k = 1}^K p\left(\Muk | \Muo, \Ro\right) \\
\propto\;&
\mathcal{W}\left(\Ro | D, (D\boldsymbol{\Sigma_a})^{-1}\right)
\prod_{k = 1}^K \mathcal{N}\left(\Muk | \Muo,  \invRo \right) \\
=\;&
\mathcal{W}(\upsilon', \boldsymbol{\Psi}')
\end{align*}
where:
\begin{align*}
\upsilon' &= D+K\\
\boldsymbol{\Psi'} &=
\left[D\boldsymbol{\Sigma_a} + \sum_k (\Muk- \Muo)(\Muk- \Muo)^T \right]^{-1}
\end{align*}

\subsubsection*{Shared base covariances $p(\Wo | \cdot)$:}
\begin{align*}
p(\Wo | \cdot) 
\propto\;&
p\left(\Wo  | D, \frac{1}{D} \boldsymbol{\Sigma_a}\right) 
\prod_{k=1}^K p\left(\Sk | \betaoa, \invWo\right)\\
\propto\;&
\mathcal{W}\left(\Wo | D, \frac{1}{D} \boldsymbol{\Sigma_a}\right)
\prod_{k=1}^K \mathcal{W}\left(\Sk | \betaoa, \left(\betaoa\Wo\right)^{-1}\right)\\
=\;&
\mathcal{W}(\upsilon', \boldsymbol{\Psi}')
\end{align*}
where:
\begin{align*} 
\upsilon' &=D + K\betaoa\\
\boldsymbol{\Psi}' &=
\left[D\boldsymbol{\Sigma_a}^{-1} +  \betaoa\sum_{k=1}^K\Sk\right]^{-1}
\end{align*}

\subsubsection*{Shared base degrees of freedom $p(\betaoa | \cdot)$:}
\begin{align*}
p(\betaoa | \cdot) 
&\propto 
p(\betaoa) \prod_{k=1}^K p\left(\Sk | \Wo ,  \betaoa\right)\\
&=p(\betaoa | 1, \frac{1}{D})\prod_{k=1}^K  \mathcal{W} \left(\Sk | \Wo ,  \betaoa\right)
\end{align*}
where there is no conjugacy we can exploit. We may sample from this distribution with Adaptive Rejection Sampling.

\section{Conditionals for the behavior view}
In this appendix we provide the conditional distributions for the behavior view to be plugged into the Gibbs sampler. Except for $\beta_{b_0}$, conjugacy can be exploited in every case and therefore their derivations straightforward and well known. The derivation for $\beta_{b_0}$ is left for another section:
\subsection{Users parameters}
\subsubsection*{Users latent coefficient $p(b_u | \cdot )$:}

Let $\mathbf{Z}$ be a $K\times U$ a binary matrix where $\mathbf{Z}_{k,u}=1$ denotes whether user $u$ is assigned to cluster $k$. Let $\mathbf{I_{[T]}}$ and $\mathbf{I_{[U]}}$ identity matrices of sizes $T$ and $U$, respectively. Let $\boldsymbol{\mu}^\text{(f)} = (\mu_1^{\text{(f)}},...,\mu_K^{\text{(f)}})$ and  $\mathbf{s}^\text{(f)} = (s_1^{\text{(f)}},...,s_K^{\text{(f)}})$  Then:
\begin{align*}
p(\mathbf{b} | \cdot ) 
\propto&\;
p(\mathbf{b} | \boldsymbol{\mu}^\text{(f)}, \mathbf{s}^\text{(f)}, \mathbf{Z}) p(\mathbf{y} | \mathbf{P, b})\\
\propto&\; 
\mathcal{N}(\mathbf{b} | \mathbf{Z}^T\boldsymbol{\mu}^\text{(f)}, \mathbf{Z}^T \mathbf{s}^\text{(f)} \mathbf{I_{[U]}}) 
\mathcal{N}(\mathbf{y}|\mathbf{P}^T \mathbf{b, \sigma_y I_{[T]}})\\
=&\; 
\mathcal{N}(\mathbf{\boldsymbol{\mu'}, \boldsymbol{\Lambda'}^{-1}})
\end{align*}
where:
\begin{align*}
\mathbf{\Lambda'} &= \mathbf{Z}^T \mathbf{s}^\text{(f)} \mathbf{I_{[U]}} +  \mathbf{P}\sigma_\textbf{y}^{-2} \mathbf{I}_{[T]} \mathbf{P}^T  \\
\boldsymbol{\mu'} &= \mathbf{\Lambda'}^{-1}(\mathbf{Z}^T \mathbf{s}^\text{(f)} \mathbf{Z}^T\boldsymbol{\mu}^\text{(f)}+ \mathbf{P} \sigma_\text{y}^{-2} \mathbf{I_{[T]} y})
\end{align*}
\subsection{Component parameters}
\subsubsection*{Components means $p(\muk| \cdot )$:}

\begin{align*}
p(\muk | \cdot ) 
&\propto
p\left(\muk | \muo, \invro\right) 
\prod_{u \in k} p(b_u | \muk, \sk, \mathbf{z})\\
&\propto
\mathcal{N}\left(\muk | \muo, \invro\right) 
\prod_{u \in k} \mathcal{N}(b_u | \muk, \sk)\\
&=\mathcal{N}(\boldsymbol{\mu'}, \boldsymbol{\Lambda'}^{-1})
\end{align*}
where:
\begin{align*}
\boldsymbol{\Lambda'} &= \ro + n_k \sk\\ 
\boldsymbol{\mu'} &= \boldsymbol{\Lambda'^{-1}} \left(\ro \muo + \sk \sum_{u\in k} b_u\right)
\end{align*}

\subsubsection*{Components precisions $p(\sk | \cdot )$:}
\begin{align*}
p(\sk | \cdot ) 
&\propto 
p(\sk |\betaof, \wo)
\prod_{u \in k} p(b_u | \muk, \sk, \mathbf{z})\\
&\propto
\mathcal{G}\left(\sk |\betaof, \left(\betaof\wo\right)^{-1}\right)
\prod_{u \in k} \mathcal{N}(b_u | \muk, \sk)\\
&=
\mathcal{G}(\upsilon', \psi')
\end{align*}
where:
\begin{align*}
\upsilon' &= \betaof + n_k\\
\psi' &=
\left[ \betaof \wo + \sum_{u \in k} \left(b_u - \muk\right)^2  \right]^{-1}
\end{align*}

\subsection{Shared hyper-parameters}
\subsubsection*{Shared base mean $p(\muo | \cdot)$:}
\begin{align*}
p(\muo | \cdot) 
&\propto  
p(\muo | \mu_{\hat{b}}, \sigma_{\hat{b}})
\prod_{k = 1}^K p(\muk | \muo, \ro )  \\
&\propto
\mathcal{N}(\muo | \mu_{\hat{b}}, \sigma_{\hat{b}})
\prod_{k = 1}^K\mathcal{N}\left(\muk | \muo, \invro\right)  \\
&= \mathcal{N}(\mu', \sigma'^{-2})
\end{align*}
where:
\begin{align*}
\sigma'^{-2} &= \sigma_{\hat{b}}^{-2} + K \ro\\ 
\mu' &= \sigma_{\hat{b}}^{2'} (\sigma_{\hat{b}}^{-2} 
\mu_{\hat{b}} + K \ro \overline{\muk})
\end{align*}

\subsubsection*{Shared base precision $p(\ro | \cdot)$}
\begin{align*}
p(\ro | \cdot) 
&\propto 
p(\ro | 1, \sigma_{\hat{b}}^{-2}) 
\prod_{k = 1}^K p(\muk | \muo, \ro) \\
&\propto
\mathcal{G}(\ro | 1, \sigma_{\hat{b}}^{-2}) 
\prod_{k = 1}^K \mathcal{N}\left(\muk | \muo,  \invro\right) \\
&= \mathcal{G}(\upsilon', \psi')
\end{align*}
where:
\begin{align*}
\upsilon' =& 1+K\\
\psi' =& \left[\sigma_{\hat{b}}^{-2} + \sum_{k=1}^K \left(\muk- \muo\right)^2\right]^{-1}
\end{align*}

\subsubsection*{Shared base variance $p(\wo | \cdot)$:}
\begin{align*}
p(\wo | \cdot) 
&\propto 
p(\wo | 1, \sigma_{\hat{b}})
\prod_{r=1}^K p\left(\sk | \betaof, \wo\right)\\
&\propto
\mathcal{G}(\wo | 1, \sigma_{\hat{b}}) 
\prod_{k=1}^K \mathcal{G}\left(\sk | \betaof, \left(\beta\wo\right)^{-1}\right)\\
&= \mathcal{G}(\upsilon', \psi')   
\end{align*}
\begin{align*}
\upsilon' =& 1 + K\betaof\\
\psi' =& 
\left[\sigma_{\hat{b}}^{-2} +  \betaof \sum_{k=1}^K \sk\right]^{-1}
\end{align*}

\subsubsection*{Shared base degrees of freedom $p(\betaof | \cdot)$:}
\begin{align*}
p(\betaof | \cdot) 
&\propto
p(\betaof) \prod_{r=1}^K p(\sk | \wo ,  \betaof)\\
&=p(\betaof | 1, 1)\prod_{r=1}^K  \mathcal{G} \left(\sk |\betaof, \left(\betaof  \wo \right)^{-1}\right)
\end{align*}
where there is no conjugacy we can exploit. We will sample from this distribution with Adaptive Rejection Sampling.

\subsection{Regression noise}
Let the precision $s_{\text{y}}$ be the inverse of the variance $\sigma_{\text{y}}^{2}$. Then:
\begin{align*}
p(s_{\text{y}} | \cdot) &\propto
p(s_{\text{y}} | 1,\sigma_{0}^{-2}) \prod_{t=1}^T p(y_t | \mathbf{p^T b}, s_{\text{y}})\\
&\propto \mathcal{G}(s_{\text{y}} | 1,\sigma_{\text{0}}^{-2}) \prod_{t=1}^T \mathcal{N}( y_t | \mathbf{p^T b}, s_{\text{y}})\\
&= \mathcal{G}(\upsilon', \psi')  
\end{align*}
\begin{align*}
\upsilon' &= 1+T\\ 
\psi' &= \left[\sigma_{\text{0}}^{2} + \sum_{t=1}^{T}\left(y_t-\mathbf{p^Tb}\right)^2\right]^{-1}
\end{align*}

\section{Sampling $\betaoa$}
For the feature view, if:
\begin{align*}
\frac{1}{\beta - D + 1} \sim \mathcal{G}(1, \frac{1}{D})
\end{align*}
we can get the prior distribution of $\beta$ by variable transformation:
\begin{align*}
p(\beta) =\;& \mathcal{G}(\frac{1}{\beta-D+1})|\frac{\partial}{\partial \beta}\frac{1}{\beta-D+1}|\\
&\propto \left(\frac{1}{\beta-D+1}\right)^{-1/2} \exp\left(-\frac{D}{2(\beta-D+1)}\right)
\frac{1}{(\beta-D+1)^2}\\
&\propto \left(\frac{1}{\beta-D+1}\right)^{3/2} \exp\left(-\frac{D}{2(\beta-D+1)}\right)
\end{align*}
Then:
\begin{align*}
p(\beta) &\propto  (\beta - D + 1)^{-3/2} \exp\left( -\frac{D}{2(\beta - D +1)}\right)
\end{align*}
The Wishart likelihood is:
\begin{align*}
\mathcal{W}(\mathbf{S}_k | \beta, (\beta\mathbf{W})^{-1})
=&
\frac{(|\mathbf{W}| (\beta/2)^D)^{\beta/2}}{\Gamma_D(\beta/2)}
|\mathbf{S}_k|^{(\beta-D-1)/2} 
\exp\left(- \frac{\beta}{2}\text{Tr}(\mathbf{S}_k\mathbf{W})\right)\\
=&
\frac{(|\mathbf{W}| (\beta/2)^D)^{\beta/2}}{\prod_{d=1}^{D} \Gamma(\frac{\beta+d-D}{2})}
|\mathbf{S}_k|^{(\beta-D-1)/2}
\exp\left(- \frac{\beta}{2}\text{Tr}(\mathbf{S}_k\mathbf{W})\right)\\
\end{align*}
We multiply both equations, the Wishart likelihood (its $K$ factors) and the prior, to get the posterior:
\begin{align*}
p(\beta | \cdot) =& 
 \left(\prod_{d=0}^D \Gamma (\frac{\beta}{2} + \frac{d-D}{2}) \right)^{-K}
\exp\left(-\frac{D}{2(\beta-D+1)}  \right)
(\beta-D+1)^{-3/2}
\\&\times
(\frac{\beta}{2})^{\frac{KD\beta}{2}}
\prod_{k=1}^K (|\mathbf{S}_k||\mathbf{W}|)^{\beta/2} \exp\left(-\frac{\beta}{2} \text{Tr}(\mathbf{S}_k\mathbf{W})\right)
\end{align*}
Then if $y = \ln\beta$:
\begin{align*}
p(y | \cdot) =\;& 
e^y
\left(\prod_{d=0}^D \Gamma (\frac{e^y}{2} + \frac{d-D}{2}) \right)^{-K}
\exp\left(-\frac{D}{2(e^y-D+1)}  \right)
(e^y-D+1)^{-3/2}
\\&\times
(\frac{e^y}{2})^{\frac{KDe^y}{2}}
\prod_{k=1}^K (|\mathbf{S}_k||\mathbf{W}|)^{e^y/2} \exp\left(-\frac{e^y}{2} \text{Tr}(\mathbf{S}_k\mathbf{W})\right)
\end{align*}
and its logarithm is:
\begin{align*}
\ln p(y | \cdot) 
=\;& 
y 
-K \sum_{d=0}^D \ln\Gamma (\frac{e^y}{2} + \frac{d-D}{2})
-\frac{D}{2(e^y-D+1)}
-\frac{3}{2}\ln(e^y-D+1)
\\&
+\frac{KDe^y}{2}(y - \ln2)
+\frac{e^y}{2} \sum_{k=1}^K \left( \ln (|\mathbf{S}_k||\mathbf{W}|) - \text{Tr}(\mathbf{S}_k\mathbf{W})\right)
\end{align*}
which is a concave function and therefore we can use Adaptive Rejection Sampling (ARS). ARS sampling works with the derivative of the log function:
\begin{align*}
\frac{\partial}{\partial y} \ln p(y | \cdot) 
=& 
1-K \frac{e^y}{2} \sum_{d=1}^D \Psi (\frac{e^y}{2} + \frac{d-D}{2})
+\frac{De^y}{2(e^y-D+1)^2}
-\frac{3}{2}\frac{e^y}{e^y-D+1}
\\&
+\frac{KDe^y}{2}(y - \ln2) + \frac{KDe^y}{2}
+\frac{e^y}{2} \sum_{k=1}^K \left(\ln (|\mathbf{S}_k||\mathbf{W}|) - \text{Tr}(\mathbf{S}_k\mathbf{W})\right)
\end{align*}
where $\Psi(x)$ is the digamma function.


\section{Sampling $\betaof$}
For the behavior view, if 
\begin{align*}
	\frac{1}{\beta} \sim \mathcal{G}(1,1)
\end{align*}
the posterior of $\beta$ is:
\begin{align*}
	p(\beta | \cdot) 
	=&  
	\Gamma(\frac{\beta}{2})^{-K}\exp\left(\frac{-1}{2\beta}\right) \left(\frac{\beta}{2}\right)
	^{(K \beta -3)/2}
	\prod_{k=1}^{K} (s_k w)^{\beta/2} \exp\left(-\frac{\beta s_k w}{2}\right)
\end{align*}
Then if $y=\ln\beta$:
\begin{align*}
	p(y | \cdot) =&  e^y \Gamma(\frac{e^y}{2})^{-K}\exp\left(\frac{-1}{2e^y}\right) \left(\frac{e^y}{2}\right)
	^{(K e^y -3)/2}
	\prod_{k=1}^{K} (s_k w)^{e^y/2} \exp\left(-\frac{e^y s_k w}{2}\right)
\end{align*}
and its logarithm:
\begin{align*}
	\ln p(y | \cdot) =& y -K\ln\Gamma \left(\frac{e^y}{2}\right) + \left(\frac{-1}{2e^y}\right)
	+\frac{Ke^y-3}{2}\left(y - \ln2\right)
	+ \frac{e^y}{2}\sum_{k=1}^{K} \left(\ln (s_k w) - s_k w \right)
\end{align*}
which is a concave function and therefore we can use Adaptive Rejection Sampling. The derivative is:
\begin{align*}
	\frac{\partial}{\partial y} \ln p(y | \cdot) =& 
	1 
	-K \Psi \left(\frac{e^y}{2}\right) \frac{e^y}{2}
	+ \left(\frac{1}{2e^y}\right)
	+\frac{Ke^y}{2} \left(y - \ln2\right) + \frac{Ke^y-3}{2}
	+ \frac{e^y}{2}\sum_{k=1}^{K}\left(\ln (s_k w) - s_k w\right)
\end{align*}
where $\Psi(x)$ is the digamma function.

\section{Sampling $\alpha$}
Since the inverse of the concentration parameter $\alpha$ is given a Gamma prior
\begin{align*}
	\frac{1}{\alpha} \sim \mathcal{G}(1,1)
\end{align*}
we can get the prior over $\alpha$ by variable transformation:
\begin{align*}
p(\alpha) 
\propto~& 
\alpha^{-3/2} \exp \left(-1/(2\alpha)\right)
\end{align*}
Multiplying the prior of $\alpha$ by its likelihood we get the posterior:
\begin{align*}
	p(\alpha | \cdot) 
	\propto~& 
	\alpha^{-3/2} \exp \left(-1/(2\alpha)\right)
	\times
	\frac{\Gamma(\alpha)}{\Gamma(\alpha+U)}
	\prod_{j=1}^{K}
	\frac{\Gamma(n_j + \alpha/K)}{\alpha/K}\\
	\propto~& 
	\alpha^{-3/2} \exp \left(-1/(2\alpha)\right)
	\frac{\Gamma(\alpha)}{\Gamma(\alpha+U)}\alpha^K \\
	\propto~&\alpha^{K-3/2} \exp \left(-1/(2\alpha)\right)
	\frac{\Gamma(\alpha)}{\Gamma(\alpha+U)}
\end{align*}
Then if $y=\ln \alpha$:
\begin{align*}
p(y | \cdot) = e^{y(K-3/2)}
\exp(-1/(2e^y))
\frac{\Gamma(e^y)}{\Gamma(e^y +U)}
\end{align*}
and its logarithm is:
\begin{align*}
\ln p(y | \cdot) = 
y(K-3/2)
-1/(2e^y)+
\ln\Gamma(e^y) - \ln\Gamma(e^y+U)
\end{align*}
which is a concave function and therefore we can use Adaptive Rejection Sampling. The derivative is:
\begin{align*}
\frac{\partial}{\partial y} \ln p(y | \cdot) = 
(K-3/2)
+1/(2e^y)+
e^y\Psi(e^y) - e^y\Psi(e^y+U)
\end{align*}

\bibliographystyle{spbasic} 
\bibliography{library}
\end{document}